\documentclass[10pt,twocolumn,letterpaper]{article}

\usepackage[pagenumbers]{cvpr} 

\makeatletter
\@namedef{ver@everyshi.sty}{}
\makeatother

\usepackage{graphicx}
\usepackage{amsmath}
\usepackage{amssymb}
\usepackage{booktabs}
\usepackage{xspace}
\usepackage{xcolor}
\usepackage{multirow}
\usepackage{makecell}
\usepackage{diagbox}
\usepackage[accsupp]{axessibility}

\usepackage[utf8]{inputenc}
\usepackage{pgfplots}
\DeclareUnicodeCharacter{2212}{−}
\usepgfplotslibrary{groupplots,dateplot}
\usetikzlibrary{patterns,shapes.arrows}
\pgfplotsset{compat=newest}

\newcommand{\neurips}{Oracle }
\newcommand{\aaai}{ZOC }
\newcommand{\actionclipbaseline}{ActionCLIP }

\definecolor{autolabelcolor}{RGB}{103, 49, 71}
\definecolor{forestgreen}{RGB}{34, 139, 34}
\definecolor{maroon}{RGB}{128, 0, 0}
\definecolor{skyblue}{RGB}{135, 206, 235}
\definecolor{cayenne}{RGB}{218, 74, 82}
\definecolor{ice}{RGB}{165, 242, 243}

\newcommand{\autolabel}{\textcolor{autolabelcolor}{\texttt{AutoLabel}}\xspace}
\newcommand{\task}{OUVDA\xspace}
\newcommand{\promptlabel}[2]{\textcolor{#1}{#2}}

\newcommand{\data}{\mathcal{D}}
\newcommand{\tsource}{\mathtt{S}}
\newcommand{\ttarget}{\mathtt{T}}
\newcommand{\tunknown}{\mathtt{U}}
\newcommand{\tcand}{\text{cand},\mathtt{T}}
\newcommand{\tpriv}{\text{priv},\mathtt{T}}
\newcommand{\R}{\mathbb{R}}
\newcommand{\E}{\mathbb{E}}

\newcommand{\supp}{{{\bf Supp Mat}\xspace}}

\newcommand{\bfv}{{\bf v}}
\newcommand{\bfw}{{\bf w}}
\newcommand{\bfx}{{\bf x}}

\newcommand{\bfX}{{\bf X}}

\def\ie{\emph{i.e.,}\xspace}
\def\eg{\emph{e.g.,}\xspace}

\def\etal{\emph{et al.}}
\def\wrt{w.r.t.\xspace}

\newcommand{\ths}{\textsuperscript{th}\;}

\usepackage{algorithm2e}
\RestyleAlgo{ruled}
\SetKwInput{KwInput}{Input}
\SetKwInput{KwOutput}{Output}
\SetKwComment{Comment}{/* }{ */}
\SetKwFunction{filter}{filter}
\SetKwProg{KwProc}{Procedure}{}{}

\usepackage[pagebackref,breaklinks,colorlinks]{hyperref}

\usepackage[capitalize]{cleveref}
\crefname{section}{Sec.}{Secs.}
\Crefname{section}{Section}{Sections}
\Crefname{table}{Table}{Tables}
\crefname{table}{Tab.}{Tabs.}

\def\cvprPaperID{****} 
\def\confName{CVPR}
\def\confYear{2023}

\begin{document}

\title{AutoLabel: CLIP-based framework for Open-set Video Domain Adaptation}

\author{
    Giacomo Zara$^{1}$, Subhankar Roy$^{3}$, Paolo Rota$^{1}$, Elisa Ricci$^{1,2}$ \\
    $^{1}$University of Trento, Italy~~$^{2}$Fondazione Bruno Kessler, Italy \\
    $^{3}$LTCI, Télécom Paris, Institut polytechnique de Paris, France \\
    \tt\small{\{giacomo.zara,paolo.rota,e.ricci\}@unitn.it, subhankar.roy@telecom-paris.fr}
}
\maketitle

\begin{abstract}
   Open-set Unsupervised Video Domain Adaptation (\task) deals with the task of adapting an action recognition model from a labelled source domain to an unlabelled target domain that contains ``target-private'' categories, which are present in the target but absent in the source. In this work we deviate from the prior work of training a specialized open-set classifier or weighted adversarial learning by proposing to use pre-trained Language and Vision Models (CLIP). The CLIP is well suited for \task due to its rich representation and the zero-shot recognition capabilities. However, rejecting target-private instances with the CLIP's zero-shot protocol requires oracle knowledge about the target-private label names. To circumvent the impossibility of the knowledge of label names, we propose \autolabel that automatically discovers and generates object-centric compositional candidate target-private class names. Despite its simplicity, we show that CLIP when equipped with \autolabel can satisfactorily reject the target-private instances, thereby facilitating better alignment between the shared classes of the two domains. The code is available\footnote{\url{https://github.com/gzaraunitn/autolabel}}.
\end{abstract}


\vspace{-6mm}
\section{Introduction}
\label{sec:intro}

Recognizing actions in video sequences is an important task in the field of computer vision, which finds a wide range of applications in human-robot interaction, sports, surveillance, and anomalous event detection, among others. Due to its high importance in numerous practical applications, action recognition has been heavily addressed using deep learning techniques~\cite{zhu2020comprehensive,kong2022human,pareek2021survey}. Much of the success in action recognition have noticeably been achieved in the supervised learning regime~\cite{simonyan2014two,feichtenhofer2017spatiotemporal,carreira2017quo}, and more recently shown to be promising in the unsupervised regime~\cite{lee2017unsupervised,xu2019self,gan2018geometry} as well. As constructing large scale annotated and curated action recognition datasets is both challenging and expensive, focus has shifted towards adapting a model from a \textit{source} domain, having a labelled source dataset, to an unlabelled \textit{target} domain of interest. However, due to the discrepancy (or \textit{domain shift}) between the source and target domains, naive usage of a source trained model in the target domain leads to sub-optimal performance~\cite{torralba2011unbiased}.

To counter the domain shift and and improve the transfer of knowledge from a labelled source dataset to an unlabelled target dataset, unsupervised video domain adaptation (UVDA) methods~\cite{chen2019temporal,choi2020shuffle,munro2020multi} have been proposed in the literature. Most of the prior literature in UVDA are designed with the assumption that the label space in the source and target domain are identical. This is a very strict assumption, which can easily become void in practice, as the target domain may contain samples from action categories that are not present in the source dataset~\cite{panareda2017open}. In order to make UVDA methods more useful for practical settings, open-set unsupervised video domain adaptation (\task) methods have recently been proposed~\cite{chen2021conditional,busto2018open}. The main task in \task comprise in promoting the adaptation between the \textit{shared} (or \textit{known}) classes of the two domains by excluding the action categories that are exclusive to the target domain, also called as \textit{target-private} (or \textit{unknown}) classes. Existing \task prior arts either train a specialized open-set classifier~\cite{busto2018open} or weighted adversarial learning strategy~\cite{chen2021conditional} to exclude the target-private classes.

\begin{figure*}[t]
  \centering
  \begin{subfigure}{0.48\linewidth}
    \includegraphics[width=\columnwidth]{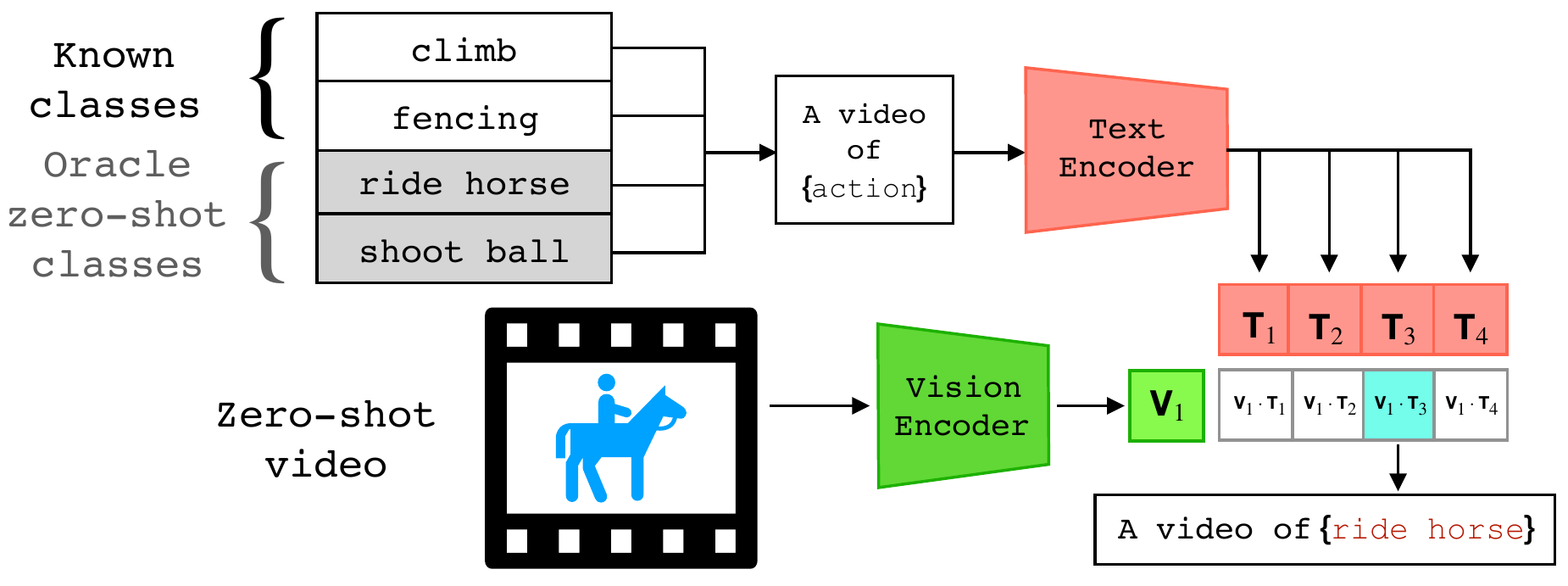}
    \caption{Zero-shot prediction using CLIP}
    \label{fig:clip-zero-shot}
  \end{subfigure}
  \hfill
  \begin{subfigure}{0.48\linewidth}
    \includegraphics[width=\columnwidth]{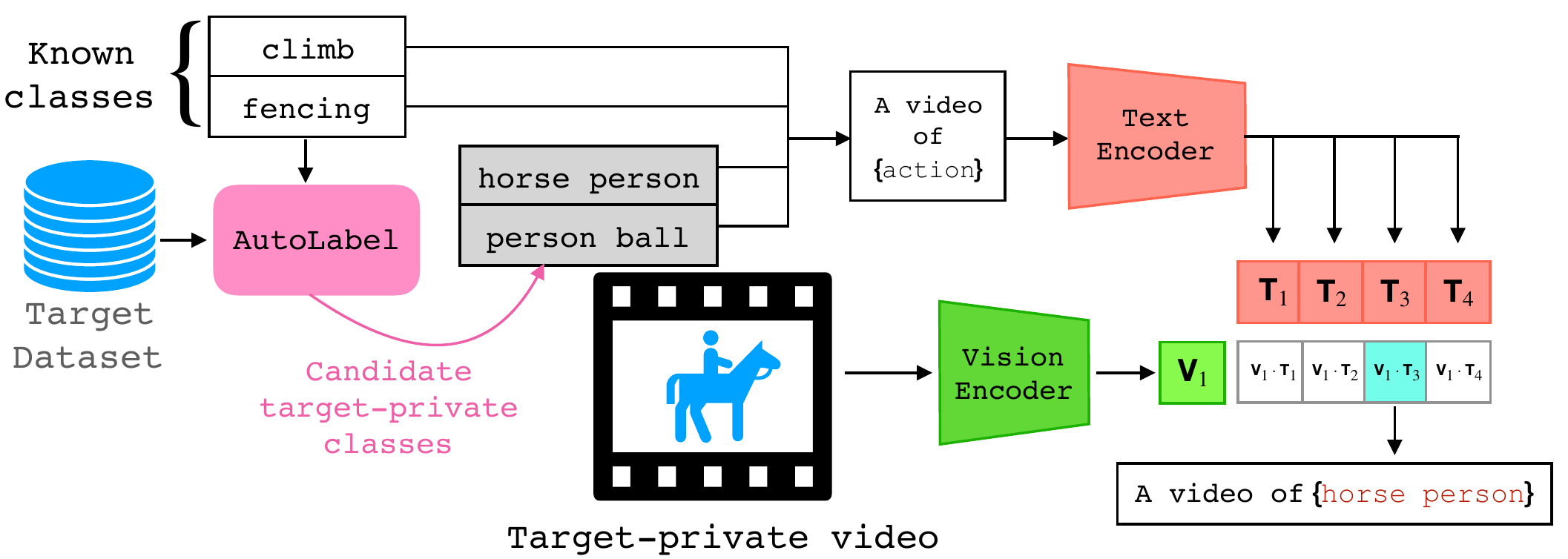}
    \caption{Rejecting target-private instances with our proposed \autolabel}
    \label{fig:autolabel}
  \end{subfigure}
  \vspace{-2mm}
  \caption{Comparison of \autolabel with CLIP~\cite{radford2021learning} for zero-shot prediction on target-private instances. (a) CLIP assumes the knowledge about the \textit{oracle} zero-shot classes names (\texttt{ride horse, shoot ball}); (b) Our proposed \autolabel discovers automatically the \textit{candidate} target-private classes (\texttt{horse person, person ball}) and extends the known classes label set}
  \label{fig:short}
  \vspace{-6mm}
\end{figure*}

Contrarily, we address \task by tapping into the very rich representations of the open-sourced foundation \textit{Language and Vision Models} (LVMs). In particular, we use CLIP (Contrastive Language-Image Pre-training)~\cite{radford2021learning}, a foundation model that is trained on \textit{web-scale} image-text pairs, as the core element of our framework. We argue that the LVMs (\eg CLIP) \textit{naturally} lend themselves well to \task setting due to: \emph{(i)} the representation learned by LVMs from \textit{webly} supervised \textit{image-caption} pairs comes encoded with an immense amount of prior about the real-world, which is (un)surprisingly beneficial in narrowing the shift in data distributions, even for video data; \emph{(ii)} the zero-shot recognition capability of such models facilitates identification and separation of the target-private classes from the shared ones, which in turn ensures better alignment between the known classes of the two domains. 

Zero-shot inference using CLIP requires multi-modal inputs, \ie a test video and a set of \textit{all} possible prompts ``\texttt{A video of \{\textcolor{blue}{label}\}}'', where \texttt{\textcolor{blue}{label}} is a class name, for computing the cosine similarity (see \cref{fig:clip-zero-shot}). However in the \task scenario, except the shared classes, \textit{a priori} knowledge about the target-private classes label names are not available (the target dataset being unlabelled). Thus, exploiting zero-shot capability of CLIP to identify the target-private instances in an unconstrained \task scenario becomes a bottleneck. To overcome this issue we propose \autolabel, an automatic labelling framework that constructs a set of \textit{candidate} target-private class names, which are then used by CLIP to potentially identify the target-private instances in a zero-shot manner.

In details, the goal of \autolabel is to augment the set of shared class names (available from the source dataset) with a set of candidate target-private class names that best represent the true target-private class names in the target dataset at hand (see \cref{fig:autolabel}). To this end, we use an external pre-trained \textit{image captioning} model ViLT~\cite{kim2021vilt} to extract a set of \textit{attribute} names from every frame in a video sequence (see~\cref{sec:attributes} for details). This is motivated by the fact that actions are often described by the constituent objects and actors in a video sequence. As an example, a video with the prompt ``\texttt{A video of \{\textcolor{blue}{chopping onion}\}}'' can be \textit{loosely} described by the proxy prompt ``\texttt{A video of \{\promptlabel{forestgreen}{knife}\}, \{\promptlabel{maroon}{onion}\} and \{\promptlabel{skyblue}{arm}\}}''
crafted from the predicted attribute names. In other words, the attributes ``\texttt{\promptlabel{forestgreen}{knife}}'', ``\texttt{\promptlabel{maroon}{onion}}'' and ``\texttt{\promptlabel{skyblue}{arm}}'' when presented to CLIP in a prompt can elicit similar response as the true action label ``\texttt{\promptlabel{blue}{chopping onion}}''. 

Naively expanding the label set using ViLT predicted attributes can introduce redundancy because: \emph{(i)} ViLT predicts attributes per frame and thus, there can be a lot of distractor object attributes in a video sequence; and \emph{(ii)} ViLT predicted attributes for the shared target instances will be duplicates of the true source action labels. Redundancy in the shared class names will lead to ambiguity in target-private instance rejection.

Our proposed framework \autolabel reduces the redundancy in the effective label set in the following manner. First, it uses unsupervised clustering (\eg k-means~\cite{lloyd1982least}) on the target dataset to cluster the target samples, and then constructs the top-\textit{k} most frequently occurring attributes among the target samples that are assigned to each cluster. This step gets rid of the long-tailed set of attributes, which are inconsequential for predicting an action (see \ref{sec:discover} for details). Second, \autolabel removes the duplicate sets of attributes that bear resemblance with the source class names (being the same shared underlying class) by using a set \textit{matching} technique. At the end of this step, the effective label set comprises the shared class names and the candidate sets of attribute names that represent the target-private class names (see \cref{sec:matching} for details). Thus, \autolabel unlocks the zero-shot potential of the CLIP, which is very beneficial in unconstrained \task.

Finally, to transfer knowledge from the source to the target dataset, we adopt conditional alignment using a simple \textit{pseudo-labelling} mechanism. In details, we provide to the CLIP-based encoder the target samples and the extended label set containing the shared and candidate target-private classes. Then we take the top-\textit{k} pseudo-labelled samples for each predicted class and use them for optimizing a supervised loss (see \cref{sec:pseudo-labels} for details). Unlike many open-set methods~\cite{chen2021conditional,busto2018open} that reject all the target-private into a single unknown category, \autolabel allows us to discriminate even among the target-private classes. Thus, the novelty of our \autolabel lies not only in facilitating the rejection of target-private classes from the shared ones, but also opens doors to open world recognition~\cite{bendale2015towards}.

In summary, our contributions are: \emph{(i)} We demonstrate that the LVMs like CLIP can be harnessed to address \task, which can be excellent replacement to complicated alignment strategies; \emph{(ii)} We propose \autolabel, an automatic labelling framework that discovers candidate target-private classes names in order to promote better separation of shared and target-private instances; and \emph{(iii)} We conduct thorough experimental evaluation on multiple benchmarks and surpass the existing \task state-of-the-art methods.

\vspace{-3mm}
\section{Related Work}
\label{sec:related}

\noindent{\bf Action Recognition.} A plethora of deep learning methods have been proposed for action recogniton, which can be roughly categorized based upon the type of network architecture used to process the video clips and the input modalities (see~\cite{kong2022human} for an extensive survey). The most common methods rely on 2D-CNNs~\cite{karpathy2014large,wang2016temporal} coupled with frames aggregation methods~\cite{girdhar2017actionvlad,kar2017adascan,donahue2015long}, and on 3D-CNNs~\cite{tran2015learning,ji20123d,carreira2017quo,hara2017learning}. Furthemore, the task has been addressed with two-stream methods introducing optical flow~\cite{simonyan2014two,feichtenhofer2016convolutional} and more recently with transformer-based architectures~\cite{girdhar2019video,lu2019learning,bulat2021space}. Despite the impressive success, these models rely on large annotated datasets to train, which is indeed a bottleneck when no annotations are available. Our work focuses on adapting a model to an unlabelled target domain by exploiting the knowledge from a labelled source domain.


\vspace{1mm}

\noindent{\bf Open-set Unsupervised Domain Adaptation.} Mitigating domain-shift with unsupervised domain adaptation methods has been an active area of research for both images (see survey in ~\cite{csurka2017comprehensive}) and videos~\cite{busto2018open,chen2021conditional,chen2019temporal,munro2020multi,choi2020shuffle,da2022dual}. In particular, closed-set video domain adaptation has been realized using the pretext task of clip order prediction~\cite{choi2020shuffle}, adversarial learning coupled with temporal attention~\cite{chen2019temporal,luo2020adversarial} and contrastive learning with multi-stream networks~\cite{kim2021learning,munro2020multi}, among others. Closed-set domain adaptation being unrealistic, adaptation methods have been proposed to address the open-set adaptation scenario, but are mostly limited to the image classification task (see survey in~\cite{csurka2022visual}). However, for \task the prior work~\cite{busto2018open,chen2021conditional} is quite scarce. Busto \etal~\cite{busto2018open} proposed a method that first learns a mapping from the source to the target. After the transformation is learned, linear one-vs-one SVMs are used to separate the unknown classes. CEVT~\cite{chen2021conditional} tackles \task by modelling the entropy of the target samples as generalised extreme value distribution, with the target-private samples lying at the tail of the distribution. The entropy is then used as weights to conduct weighted adversarial alignment. Differently, our work leans on the excellent zero-shot capability of CLIP~\cite{radford2021learning} to detect the target-private instances. Post detection, we resort to conditional alignment to align the shared classes in the source and target domains using pseudo-labelling.

\vspace{1mm}

\noindent{\bf Language and Vision Models.} Off late, the vision community has witnessed a paradigm shift with the advent of the language and vision (LVM) foundation models~\cite{bommasani2021opportunities}. LVMs derive strength from the large scale pre-training with web-scale multi-modal image-text or image-audio training pairs. In particular, CLIP~\cite{radford2021learning} has demonstrated excellent downstream performance on zero-shot classification. Inspired by its success, CLIP has been adapted as ActionCLIP~\cite{wang2021actionclip} and VideoCLIP~\cite{xu2021videoclip} for addressing zero-shot action recognition and action segmentation, respectively. While ActionCLIP assumes knowledge about the oracle zero-shot classes, VideoCLIP uses a thresholded closed-set predictor to identify the unknown classes. We argue that both are impractical and sub-optimal, and therefore propose to discover the target-private classes in an automatic manner. Very recently, Esmaeilpour \etal~\cite{esmaeilpour2022zero} indeed proposed to detect zero-shot classes without any oracle knowledge, but limit themselves to image classification task only. In contrast, our \autolabel takes additional care in reducing redundancy and is thought-out for the action recognition task. 
\vspace{-2mm}
\section{Methods}
\label{sec:method}

\begin{figure*}[!t]
    \centering
    \includegraphics[width=\textwidth]{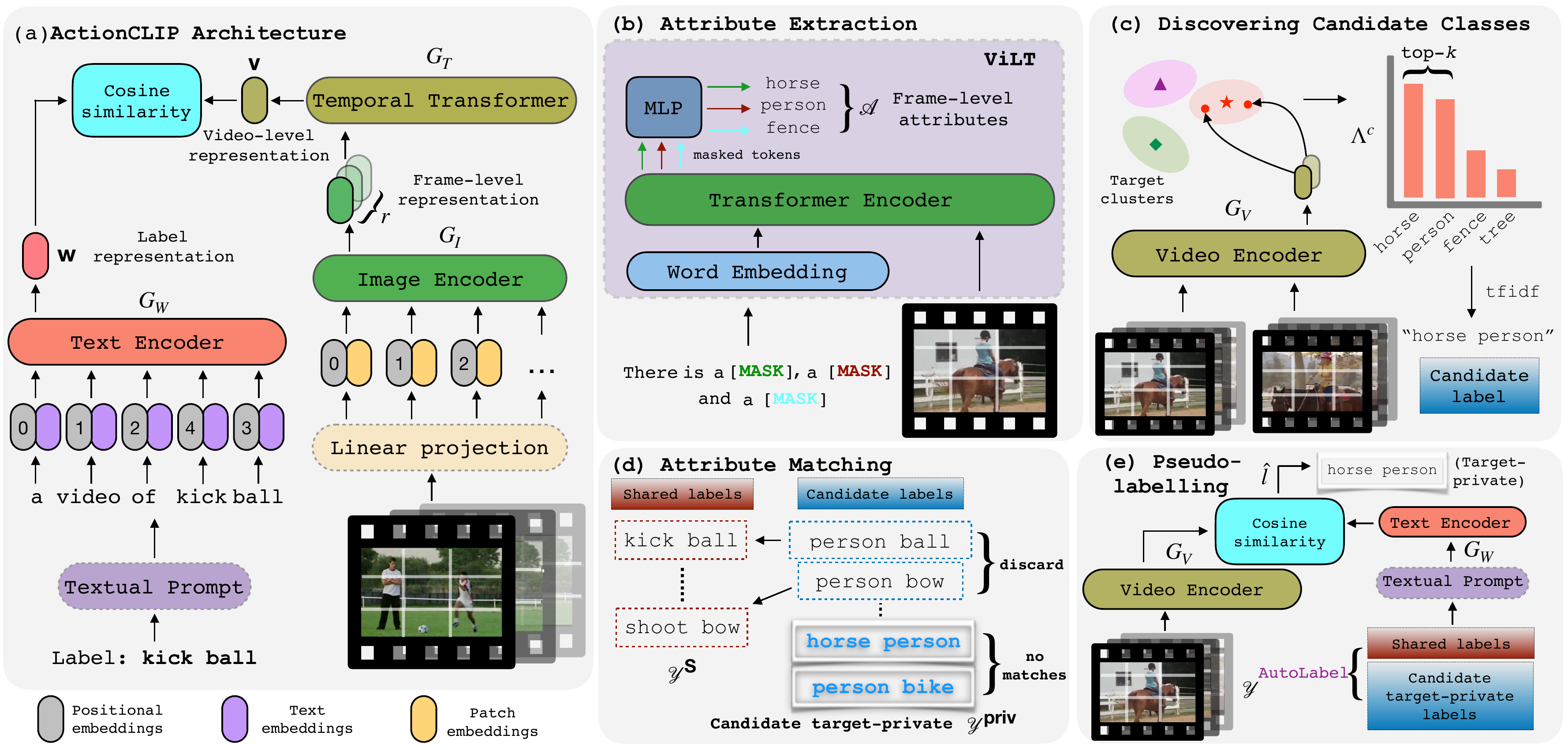}
    \caption{Overview of the \autolabel framework for \task: (a) the ActionCLIP~\cite{wang2021actionclip} forms the backbone of \autolabel for extracting video-level and text label representations; (b) per-frame attributes are extracted using ViLT~\cite{kim2021vilt}; (c) candidate class names in the target dataset are discovered in an unsupervised manner; (d) redundant candidate labels are discarded and candidate target-private class names are identified with attribute matching; and (e) extended label set $\mathcal{Y}^{\autolabel}$ enables the rejection of target-private instances} 
    \label{fig:autolabel-pipeline}
    \vspace{-5mm}
\end{figure*}

In this work we propose \autolabel to tackle the task of adapting a model from a labelled source dataset to an unlabelled target dataset, under the constraint that the target contains samples from action categories that are not present in the source domain. Before we describe the details of \autolabel we formalize \task and present the preliminaries used in our framework.

\vspace{1mm}

\noindent{\bf Problem Definition and Notations.}
\label{sec:definition}
Let us assume that we are given a source dataset containing \textit{labelled} video sequences $\data^\tsource = \{(\bfX^\tsource_i, l^\tsource_i)\}^n_{i=1}$, where $\bfX \in \mathcal{X}$ represents the input video and $l^\tsource \in \mathcal{Y}^\tsource = \{l_1, l_2, \dots, l_{K}\}$ being the $K$ \textit{shared} class names, instead of class indices. For \eg $\mathcal{Y}^\tsource = \{``\texttt{climb}", ``\texttt{fencing}", \dots, ``\texttt{push up}"\}$. Moreover, we are also given an \textit{unlabelled} target dataset $\data^\ttarget = \{\bfX^\ttarget_i\}^n_{i=1}$ containing $n$ samples from classes $\mathcal{Y}^\ttarget$. In particular, $\mathcal{Y}^\ttarget = \mathcal{Y}^\tsource \cup \mathcal{Y}^\tunknown$, where $\mathcal{Y}^\tunknown = \{l_{K+1}, l_{K+1}, \dots, l_{K+M}\}$ represents the $M$ \textit{target-private} class names and are not known to us \textit{a priori}. Each video sequence is composed of $r$ frames $\bfX = \{\bfx_j\}^r_{j=1}$ depicting an action with label $l$.

The goal in \task is to learn a parameterized function $f_\theta \colon \mathcal{X} \to \mathcal{Y}$ using $\data^\tsource \cup \data^\ttarget$, that can correctly predict the shared target instances to be belonging to one of the classes in $\mathcal{Y}^\tsource$ and reject the target-private instances as ``unknown''.

\vspace{1mm}

\noindent{\bf Overview.}
\label{sec:overview}
To address \task we propose \autolabel (see \cref{fig:autolabel-pipeline}), a CLIP-based framework, that comprise of a transformer~\cite{vaswani2017attention} as the text encoder and ViT~\cite{dosovitskiy2020image} as the vision encoder. In practice, we use the ActionCLIP architecture~\cite{wang2021actionclip} that has an additional self-attention temporal pooling to aggregate the frame-level features to output video-level feature. To enable the identification of target-private instances without any access to $\mathcal{Y}^\tunknown$, \autolabel generates a set of candidate target-private classes names. 

In details, \autolabel uses pre-trained ViLT~\cite{kim2021vilt} to predict a set of attributes from the frames in target video sequences. Then it uses unsupervised clustering on the video-level target features to cluster the video sequences into respective semantic clusters. All the predicted frame-level attributes, pertaining to the video sequences that are assigned to a given cluster, are filtered to yield the top-\textit{k} most salient attributes. These top-\textit{k} attributes are then concatenated to form a \textit{proxy} action label name. Note that this step will produce one candidate label name per cluster, including the ones corresponding to the shared classes. To disambiguate and merge the redundant class label names with the known shared class names we use set matching. At the end of this step the effective label set will comprise of the shared class names and the candidate target-private class names.

Following the extended label set creation, pseudo-labels are computed for the unlabelled target samples by providing to the text and vision encoders of ActionCLIP, as input, the target video sequences and the extended label set, respectively. The top-\textit{k\%} most confident predictions from each predicted class (computed using \texttt{argmax} on the output probability distribution) are then used as hard pseudo-labels for optimizing the ActionCLIP multi-modal training objective. Below we summarize the ActionCLIP.

\vspace{-3.2mm}

\subsection{Preliminaries}
\label{sec:prelim}

ActionCLIP~\cite{wang2021actionclip} is a transformer-based multi-modal supervised action recognition framework that consist of a text encoder $G_W(\cdot)$ to extract features of text labels and a video encoder $G_V(\cdot)$ to extract spatio-temporal features from a corresponding video (see~\cref{fig:autolabel-pipeline}a). The video encoder consists of an image encoder $G_I(\cdot)$, operating at frame-level, and a temporal transformer $G_T(\cdot)$ for aggregation, such that $G_V = G_T \circ G_I$. ActionCLIP is trained to maximize the cosine similarity $cos(\cdot, \cdot)$ between the pairwise video and label representations, which is defined as:

\vspace{-3mm}


\begin{equation}
\label{eqn:cosine-sim}
    cos(\bfX, l) = \frac{\bfv \cdot \bfw^{\texttt{T}}}{||\bfv|| \; ||\bfw||}
\end{equation}
where $\bfv = G_V(\bfX)$ and $\bfw = G_W(l)$ are the feature representations of the video and text modalities, respectively. The softmax-normalized similarity score for the $i$\ths training pair is given as:
\vspace{-2mm}
\begin{equation}
\label{eqn:softmax-sim}
    p_l(\bfX_i, l_i) = \frac{\text{exp}(cos(\bfX_i, l_i) / \tau)}{\sum^{N}_{j=1}\text{exp}(cos(\bfX_i, l_j) / \tau)}
\end{equation}
where $N$ is the number of training pairs in a mini-batch and $\tau$ is a temperature hyperparameter. The ground-truth similarity score $q$ is 1 and 0 for positive and negative pairs, respectively. The symmetric video-text constrastive loss used to train ActionCLIP on the source dataset is defined with the Kullback-Leibler (KL) divergence as:

\vspace{-3mm}
\begin{equation}
\label{eqn:actionclip}
    \mathcal{L}_\text{ActionCLIP} = \frac{1}{2} \E_{(\bfX^\tsource, l^\tsource) \sim \data^\tsource} [\text{KL}({\bf p}|{\bf q}) + \text{KL}({\bf q}|{\bf p})]
\end{equation}

\subsection{AutoLabel Framework}
\label{sec:autolabel}

The main task of the proposed \autolabel framework is to equip the ActionCLIP with zero-shot detection capabilities such that the resulting framework can be used in \task, without any oracle knowledge about the target-private class names $\mathcal{Y}^\tunknown$. To this end, \autolabel relies on a series of four sub-modules: \emph{(i)} Attribute extraction is in charge of extracting a set of frame-level attributes (\eg actors and objects) depicted in a video frame (see~\cref{fig:autolabel-pipeline}b); \emph{(ii)} the second sub-module (see \cref{fig:autolabel-pipeline}c) uses the predicted attributes to construct a set of candidate labels for the target domain that correspond to different semantic categories; \emph{(iii)} Attribute matching sub-module (see \cref{fig:autolabel-pipeline}d) further reduces the redundancy in the candidate label names that are duplicates of the known shared class names $\mathcal{Y}^\tsource$; and \emph{(iv)} Pseudo-labelling (see \cref{fig:autolabel-pipeline}e) sub-module then uses the shared and candidate target-private class names to reject the target-private instances. Next we elaborate each sub-module of \autolabel in detail.

\vspace{-5mm}

\subsubsection{Attribute Extraction}
\label{sec:attributes}

As discussed in \cref{sec:intro}, our automatic labelling approach is motivated by the fact that actions are often times described by the \textit{objects} and \textit{actors} in a video. Thus, modelling the attributes alone and constructing the candidate label names from such attributes can suffice to reject target-private instances, which is one of the two main goals in \task. We use an off-the-shelf image captioning model ViLT~\cite{kim2021vilt} and prompt it in a way to obtain a set of attributes per frame.

Specifically, we give to the ViLT model a video frame $\bfx^\ttarget_j$ and a prompt $z = $ ``\texttt{There is a [\promptlabel{forestgreen}{{\bf MASK}}], a [\promptlabel{cayenne}{{\bf MASK}}] and a [\promptlabel{ice}{{\bf MASK}}]}''. The model outputs a set of most probable words for the $m$ masked tokens as:
\vspace{-1mm}
\begin{equation}
\label{eqn:vilt-attr}
    \mathcal{A}(\bfx^\ttarget_j) = \texttt{ViLT}(\bfx^\ttarget_j, z)
\end{equation}
where $\mathcal{A}$ denote a set of attributes with $m=$ card($\mathcal{A}$). As an example in the~\cref{fig:autolabel-pipeline}b, the $\texttt{ViLT}$ predicts $m=3$ attributes: ``\texttt{horse}'', ``\texttt{person}'' and ``\texttt{fence}'' corresponding to the three masked tokens in the prompt $z$.

In a similar spirit, ZOC~\cite{esmaeilpour2022zero} used an image captioning model to generate attributes for enabling zero-shot prediction. However, we differ from ZOC in the following ways: (i) ZOC addresses image classification whereas we tackle action recognition, (ii) ZOC treats each attribute as a candidate class, while we create compositional candidate class names by combining multiple attributes. This is crucial in action recognition because action names arise from the interaction of objects and actors; and (iii) Unlike us, ZOC does not handle the redundancy caused by duplicate candidate label and shared class names. Next we describe how \autolabel combines attributes to create candidate action label names and how such redundancies can be eliminated.

\vspace{-5mm}

\subsubsection{Discovering Candidate Classes}
\label{sec:discover}

As mentioned above, an action label name is a product of the interaction between object(s) and actor(s). For instance, the action label ``\texttt{ride horse}'' depicts an interaction between a ``\texttt{horse}'' and a ``\texttt{person}''. If the attributes in $\mathcal{A}$ are treated as candidate labels in isolation, like ZOC, then for a video of horse riding the cosine similarity between the visual and text representations of both ``\texttt{horse}'' and ``\texttt{person}'' text will be high. Moreover, there can be other distractor attributes such as ``\texttt{fence}'', which if present in a frame, will also elicit high response from CLIP. In order to uniquely model the true label ``\texttt{ride horse}'' we propose the following strategy.

First, we use the video encoder $G_V$ to cluster all the target videos of $\data^\ttarget$ into $\mathcal{C}$ target clusters, which ideally should represent semantic categories. Note that we do not assume \textit{a priori} knowledge about the true cardinality of $|\mathcal{Y}^\tunknown|$ and we set $|\mathcal{C}| > |\mathcal{Y}^\tsource|$. In details, we use the standard clustering algorithm \textit{k}-means \cite{kmeans} that takes as input a set of the video-level features $\bfv^\ttarget = G_V(\bfX^\ttarget)$ and assigns them to $\mathcal{C}$ distinct centroids $\mu_c$, with the cluster assignment for the $i$\ths video sequence given as $y_i \in \{0, 1\}^{|\mathcal{C}|}$. Next, we construct a histogram per target cluster (see~\cref{fig:autolabel-pipeline}c), by using all the attributes $\Lambda^{c, \ttarget} = \{\mathcal{A}(\bfx^\ttarget) | \bfx^\ttarget \in \bfX^\ttarget_{\hat{y}^{c, \ttarget}}, \bfX^\ttarget \in \data^\ttarget \}$ associated to a target cluster $c$, where $\hat{y}^{c, \ttarget}  = \text{arg min}_{c \in \mathcal{C}} ||\mu_c - \bfX^\ttarget||$. Note that this step is carried out at the beginning of each training epoch.

We expect the most frequent attributes associated to a cluster to be the most salient and descriptive of the action. As shown in~\cref{fig:autolabel-pipeline}c, in the ``\texttt{ride horse}'' cluster, ``\texttt{horse}'' and ``\texttt{person}'' will be the most frequent attributes, and the rest will lie in the tail of the distribution. We filter the \textit{t} most common and relevant attributes in $\Lambda^{c, \ttarget}$ to obtain $\bar{\Lambda}^{c, \ttarget} = \texttt{tfidf}(\text{argtop}_k (\Lambda^{c, \ttarget}))$. Finally, we concatenate the attributes in $\bar{\Lambda}^{c, \ttarget}$ to form the candidate label:

\vspace{-3mm}
\begin{equation}
\label{eqn:candidate-label}
    l^{\tcand}_c = \bar{\Lambda}^{c, \ttarget}_1 || \dots || \bar{\Lambda}^{c, \ttarget}_t
\end{equation}
where $\cdot || \cdot$ represent the concatenation operation separated by a space. Details about the \texttt{tfidf}$(\cdot)$ operator is provided in the \supp.  Since, the target is unlabelled, we can not yet distinguish the shared candidate labels from the target-private ones. Thus, to identify the target-private instances we need a mechanism to tell apart the target-private class names, which we describe next.


\vspace{-5mm}

\subsubsection{Attribute Matching}
\label{sec:matching}

The attribute matching step is in charge of finding the candidate label names that correspond to the target-private classes. To this end, we simply find the candidate label names in $\mathcal{Y}^{\tcand} = \{l^{\tcand}_1, \dots, l^{\tcand}_{|\mathcal{C}|}\}$ that correspond to the shared label names in the source $\mathcal{Y}^\tsource$. This will essentially leave us with the candidate target-private label names that have no match with the source (see~\cref{fig:autolabel-pipeline}d).

In details, we repeat the sub-module described in~\cref{sec:discover} on the source samples to obtain the set of attributes $\bar{\Lambda}^{l^\tsource}$, where $l^\tsource$ is a source label name. Then we create a similarity matrix $S \in \R^{K \times |\mathcal{C}|}$, where an entry $s_{i,j}$ in $S$ denotes how similar a $i$\ths source label name $l^\tsource_i$ is to a $j$\ths candidate label name $l^{\tcand}_j$, and is formally defined as:

\vspace{4mm}

\[
\label{eqn:score}
    s_{i,j} = sim({\bar{\Lambda}^{l^\tsource_i}}, {\bar{\Lambda}^{j, \ttarget}})
\]

\noindent where $sim(\cdot, \cdot)$ is a scoring function that computes a similarity score based on the common attributes between the two sets and their occurrences. More details about the $sim(\cdot, \cdot)$ can be found in the \supp. If the score is higher than a threshold $\gamma$, the two sets are considered as matched, and the $j$\ths target candidate label is discarded. After this thresholding step, the candidate labels in $\mathcal{Y}^{\tcand}$ that are not matched with any of the source labels become the candidate target-private label names, such that $\mathcal{Y}^{\tpriv} \subset \mathcal{Y}^{\tcand}$.

\vspace{-4mm}

\subsubsection{Conditional Alignment}
\label{sec:pseudo-labels}

After the attribute matching, we have an extended label set $\mathcal{Y}^{\autolabel} = \mathcal{Y}^\tsource \cup \mathcal{Y}^{\tpriv}$, which comprise of the shared labels $\mathcal{Y}^\tsource$ and the candidate target-private labels $\mathcal{Y}^{\tpriv}$. The CLIP is now equipped to detect target-private instances. Specifically, for a target sample $\bfX^\ttarget$ we compute the predicted label (see~\cref{fig:autolabel-pipeline}e) as:

\vspace{-4mm}
\begin{equation}
\label{eqn:pseudo}
    \hat{l} = \text{arg max}_{l \in \mathcal{Y}^{\autolabel}} \frac{\text{exp}(cos(\bfX^\ttarget, l) / \tau)}{\sum^{|\mathcal{Y}^{\autolabel}|}_{j=1} \text{exp}(cos(\bfX^\ttarget, l_j) / \tau)}
\end{equation}
where the $\bfX^\ttarget$ is considered as a target-private if $\hat{l} \in \mathcal{Y}^{\tpriv}$.

To align the shared classes of the two domains we resort to a simple conditional alignment strategy with the usage of pseudo-labels computed using~\cref{eqn:pseudo}. In practice, we extract the top-\textit{k\%} most confident pseudo-labels per predicted class and backpropagate the supervised ActionCLIP loss in~\cref{eqn:actionclip} for the pseudo-labelled target samples. Besides the shared classes, this loss also promotes discrimination among the target-private classes as gradients are also backpropagated corresponding to the discovered candidate target-private classes $\mathcal{Y}^{\tpriv}$. This is indeed promising from the open world recognition point of view because the target-private samples are assigned to their respective semantic categories, instead of a single ``unknown'' category.

\vspace{-2mm}

\section{Experiments}
\label{sec:exp}

\noindent\textbf{Datasets and Settings.} 
We evaluate our proposed method and the baselines on two benchmarks derived from the datasets HMDB~\cite{hmdb2011hildegard}, UCF101~\cite{ucf101khurram} and Epic-Kitchens~\cite{epic_kitchens}, for the \task task. The first \task benchmark is the \textit{HMDB$\leftrightarrow$UCF}, introduced by Chen \etal~\cite{chen2021conditional}, that comprises action categories, where six of them are shared between the two domains and the remaining six are target-private. The second benchmark is the \textit{Epic-Kitchens} (EK) that is composed of egocentric videos from three kitchen environments (or domains D1, D2 and D3). We extend the closed-set UVDA EK benchmark, used in~\cite{munro2020multi,kim2021learning}, to the \task scenario where there are eight shared action categories and all the remaining ones in each kitchen environment are considered as target-private. The statistics of the two benchmarks are provided in the \supp.


\vspace{1mm}

\noindent\textbf{Implementation details.} 
As discussed in the~\cref{sec:prelim}, we employ the network architecture from ActionCLIP~\cite{wang2021actionclip}, which is composed of transformer-based video and text encoders. In particular, the video encoder contains a ViT-B/32 vision transformer, which is a pre-trained CLIP encoder~\cite{radford2021learning}, while the temporal transformer is initialized from scratch. The label representations are obtained with a 512-wide transformer containing 8 attention heads.

We extract five attributes ($m=5$) from the pre-trained image captioning model ViLT~\cite{kim2021vilt}, and set the number of attributes $t$ in a candidate label to 5 and 2 for \textit{HMDB$\leftrightarrow$UCF} and EK, respectively. To train our framework, we used AdamW optimizer~\cite{loshchilov2017decoupled}, with a learning rate of \(5 \times 10^{-5}\) and a weight decay rate of 0.2. The models were trained for 20 epochs using a total mini-batch size of 96. Each training video sequence is composed of randomly sampled \(r = 8\) frames, each of resolution 224 $\times$ 224.

\begin{table*}[!h]
    \centering
    \small
    \def\arraystretch{.7}
        \begin{tabular}{l|c|ccc|c|ccc|c}
            \toprule
            \multirow{2}{*}{Method} & \multirow{2}{*}{Backbone} & \multicolumn{4}{c|}{\textit{HMDB$\rightarrow$UCF}} & \multicolumn{4}{c}{\textit{UCF$\rightarrow$HMDB}} \\
            & & \textbf{ALL} & $\textbf{OS}^*$ & \textbf{UNK} & \textbf{HOS} & \textbf{ALL} & $\textbf{OS}^*$ & \textbf{UNK} & \textbf{HOS} \\
            \midrule
            DANN~\cite{dann} + OSVM~\cite{osvm} & \multirow{8}{*}{ResNet101~\cite{resnet}} & 64.6 & 62.9 & 74.7 & 68.3 & 66.1 & 48.3 & 83.9 & 61.3 \\ 
            JAN~\cite{jan} + OSVM~\cite{osvm} & & 61.5 & 62.9 & 73.8 & 67.9 & 61.1 & 47.8 & 74.4 & 58.2 \\
            AdaBN~\cite{adabn} + OSVM~\cite{osvm} & & 62.9 & 58.8 & 73.3 & 65.3 & 62.9 & \underline{58.8} & 73.3 & 65.3 \\
            MCD~\cite{mcd} + OSVM~\cite{osvm} & & 66.7 & 63.5 & 73.8 & 68.3 & 66.7 & 57.8 & 75.6 & 65.5 \\
            TA$^2$N~\cite{chen2019temporal} + OSVM~\cite{osvm} & & 63.4 & 61.3 & 79.0 & 69.1 & 65.3 & 56.1 & 74.4 & 64.0 \\
            TA$^3$N~\cite{chen2019temporal} + OSVM~\cite{osvm} & & 60.6 & 58.4 & 82.5 & 68.4 & 62.2 & 53.3 & 71.7 & 61.2 \\
            OSBP~\cite{Saito_2018_ECCV} + AvgPool & & 64.8 & 55.3 & \underline{85.7} & 67.2 & 67.2 & 50.8 & 84.5 & 63.5 \\
            CEVT~\cite{chen2021conditional} & & \underline{70.6} & \underline{66.8} & 84.3 & \underline{74.5} & \underline{75.3} & 56.1 & \underline{94.5} & \underline{70.4} \\
            
            \midrule
            
            CEVT-CLIP~\cite{chen2021conditional} & \multirow{4}{*}{CLIP~\cite{radford2021learning}} & 70.9 & 68.0 & 92.5 & 78.4 & 78.3 & 61.2 & 92.1 & 73.5 \\
            
            \actionclipbaseline\cite{wang2021actionclip} & & \textbf{79.7} & 81.1 & \textbf{94.3} & 87.2 & 84.8 & 75.7 & 91.5 & 82.9 \\
            
            ActionCLIP-\aaai\cite{esmaeilpour2022zero} & & 74.7 & \textbf{84.0} & 68.5 & 75.5 & 85.4 & 75.5 & \textbf{93.2} & 83.4 \\
            
            
            
            
            \autolabel (ours) & & \textbf{79.7} & 82.5 & \textbf{94.3} & \textbf{88.0} & \textbf{86.0} & \textbf{82.9} & 88.2 & \textbf{85.5} \\
            
            \midrule
            
            ActionCLIP-\textit{Oracle} \cite{fort2021exploring} & CLIP~\cite{radford2021learning} & \textit{93.3} & \textit{93.7} & \textit{100.0} & \textit{96.7} & \textit{92.8} & \textit{86.5} & \textit{98.3} & \textit{92.0} \\
            \bottomrule
        \end{tabular}
        \vspace{-2mm}
        \caption{Comparison with the state-of-the-art on the \textit{HMDB$\leftrightarrow$UCF} benchmark for \task. The best performances of the CLIP-based and ResNet-based methods are shown in bold and underlines, respectively. The \textit{Oracle} performance is shown in italics. Using CLIP as backbone greatly improves the HOS scores. The target-private rejection with our \autolabel outperforms all the baseline methods}
    \label{tab:results_hu}
\end{table*}

\begin{table*}[t]
    \centering
    \small
    \begin{tabular}{l|cccccc|c}
        \toprule
        \multirow{2}{*}{Method} & \multicolumn{7}{c}{\textit{Epic-Kitchens}} \\
        
        & D2$\rightarrow$D1 & D3$\rightarrow$D1 & D1$\rightarrow$D2 & D3$\rightarrow$D2 & D1$\rightarrow$D3 & D2$\rightarrow$D3 & Avg \\
        \midrule
        
        CEVT~\cite{chen2021conditional} & 13.2 & 14.7 & 8.4 & 16.0 & 8.1 & 11.3 & 12.0 \\
        
        CEVT-CLIP~\cite{chen2021conditional} & 13.2 & 17.4 & 13.3 & 14.3 & 10.2 & 10.1 & 13.0 \\
        \actionclipbaseline\cite{wang2021actionclip} & 31.3 & 28.3 & 38.1 & 43.4 & 29.0 & 24.2 & 32.4 \\
        
        ActionCLIP-\aaai\cite{esmaeilpour2022zero} & 25.9 & 26.8 & 31.7 & 41.3 & 28.0 & 32.1 & 30.9 \\
        
        
        
        
        \autolabel (ours) & \textbf{34.8} & \textbf{38.3} & \textbf{44.1} & \textbf{50.4} & \textbf{31.9} & \textbf{29.8} & \textbf{38.2} \\
        
        \midrule
        
        ActionCLIP-\textit{Oracle}~\cite{fort2021exploring} & \textit{33.2} & \textit{33.1} & \textit{37.1} & \textit{44.6} & \textit{24.2} & \textit{28.9} & \textit{33.5} \\
        \bottomrule
    \end{tabular}
    \vspace{-2mm}
    \caption{Comparison with the state-of-the-art on the \textit{Epic-Kitchens} benchmark for \task. The HOS scores are reported for all the methods. The best performances are shown in bold. Overall, our \autolabel surpasses all the baselines, including the \textit{Oracle}}
    \vspace{-5mm}
    \label{tab:3_kitchens}
\end{table*}

\vspace{1mm}

\noindent\textbf{Evaluation Metrics.} 
We evaluate the performance of the models on the target dataset using standard open-set metrics, as in \cite{chen2021conditional}. In particular, we report: the \textbf{ALL} accuracy which is the percentage of correctly predicted target samples over all the target samples; the closed-set $\textbf{OS}^*$ accuracy which computes the averaged accuracy over the known classes only; the \textbf{UNK} recall metric which denotes a ratio of the number of correctly predicted unknown samples over the total number of unknown samples in the target dataset; and \textbf{HOS} is the harmonic mean between $\textbf{OS}^*$ and \textbf{UNK} metrics, \ie \textbf{HOS} = $2 \times \frac{\textbf{OS}^* \times \textbf{UNK}}{\textbf{OS}^* + \textbf{UNK}}$. As the \textbf{HOS} takes into account both the closed and open-set scores, it is considered as the most meaningful metric for evaluating open-set algorithms~\cite{ovanet,chen2021conditional,bucci2020effectiveness}.


\subsection{Comparison with the State of the Art}
\label{sec:sota}

\noindent\textbf{Baselines.} We compare our \autolabel with \textbf{CEVT}~\cite{chen2021conditional}, an existing state-of-the-art method for \task. However, note that CEVT uses ResNet-101~\cite{resnet} (pre-trained on ImageNet-1k) as a backbone for extracting frame-level features, which is weaker \wrt the CLIP pre-trained backbone used by \autolabel. To be fairly comparable, we create a baseline \textbf{CEVT}-\textbf{CLIP} that replaces the ResNet-101 backbone of CEVT with the stronger CLIP vision encoder. 

Additionally, we introduce few more baselines for \task that use the representation power of CLIP, but \textit{without} the \textit{a priori} knowledge about the true target-private label set names. These baselines differ by how the target-private instances are rejected: (i) the \textbf{ActionCLIP} baseline that rejects target-private instances by thresholding the similarity values computed using the shared class names, (ii) the \textbf{ActionCLIP}-\textbf{ZOC} baseline implements the target-private rejection mechanism of ZOC~\cite{esmaeilpour2022zero}; and (iii) the \textbf{ActionCLIP}-\textit{Oracle} that assumes the knowledge of the true target-private label set names, as described in~\cite{fort2021exploring}. Note that all these baselines then fine-tune on the pseudo-labelled target data, similar to our proposed \autolabel framework.

For the \textit{HMDB$\leftrightarrow$UCF} adaptation scenario we also include the baselines reported in~\cite{chen2021conditional}, which are closed-set~\cite{chen2019temporal,dann,jan} and open-set methods~\cite{Saito_2018_ECCV} adapted to the \task setting, accordingly.


\vspace{1mm}

\noindent\textbf{Results on \textit{HMDB$\leftrightarrow$UCF}.} In~\cref{tab:results_hu} we report the results obtained on the \textit{HMDB$\leftrightarrow$UCF} benchmark. We separately report the baselines that use ResNet101 and CLIP backbones as feature extractors. It is evident that the usage of CLIP-based backbone can significantly improve the HOS scores over the baselines that use ResNet101 as feature extractor. For \eg by changing the backbone alone, the \textbf{CEVT}-\textbf{CLIP} is +3.9 and +3.1 points better than the \textbf{CEVT} for the settings \textit{HMDB$\rightarrow$UCF} and \textit{UCF$\rightarrow$HMDB}, respectively. This highlights the importance of the LVMs in closing the domain gap for the \task task.

Next, we demonstrate that \task can further benefit from CLIP by successfully rejecting the target-private instances with our \autolabel, where we observe +9.6 and +12.0 points improvement over \textbf{CEVT}-\textbf{CLIP} on the two settings, as far as the HOS scores are concerned. Our \autolabel also surpasses the \textbf{ActionCLIP} baseline in HOS score (by +0.8 and +2.6 points), which does not model the unknown classes and uses a threshold-based target-private rejection with the closed-set classifier. Interestingly, the \textbf{ActionCLIP}-\textbf{ZOC} baseline, which indeed models the unknown classes using externally trained image-captioning model like \autolabel, exhibits inferior performance than \autolabel on both the occasions (-12.5 and -2.1 points). This emphasizes the importance of carefully modelling the candidate target-private label names for the action recognition task, which is lacking in ZOC~\cite{esmaeilpour2022zero}, as previously discussed in~\cref{sec:attributes}. 

\vspace{-0.5mm}

Finally, when the oracle target-private label set names are available, CLIP demonstrates even stronger performance, as shown in the last row of~\cref{tab:results_hu}. However, when such privileged information about target datasets is not available, our \autolabel can effectively close the gap, reaching closer to the upper bound performance.

\vspace{-0.8mm}


\begin{figure}[!t]
    \vspace{-3mm}
    \centering\scriptsize
\begin{tikzpicture}

\definecolor{darkslategray38}{RGB}{38,38,38}
\definecolor{lavender234234242}{RGB}{234,234,242}
\definecolor{lightgray204}{RGB}{204,204,204}
\definecolor{purple}{RGB}{128,0,128}

\begin{axis}[
axis background/.style={fill=lavender234234242},
axis line style={white},
legend cell align={left},
legend style={
  font=\tiny,
  fill opacity=0.8,
  draw opacity=1,
  text opacity=1,
  at={(0.97,0.02)},
  anchor=south east,
  draw=lightgray204,
  fill=white
},
title={{\tiny HMDB $\rightarrow$ UCF}},
height=0.53\linewidth,
width=0.53\linewidth,
tick align=outside,
x grid style={white},
xlabel=\textcolor{darkslategray38}{\(\displaystyle C\) (\# target clusters)},
xmajorgrids,
xmajorticks=true,
xtick pos=left,
xmin=-1, xmax=60,
xtick style={color=darkslategray38},
y grid style={white},
ylabel=\textcolor{darkslategray38}{HOS (\%)},
ymajorgrids,
ymajorticks=true,
ytick pos=left,
ymin=60, ymax=95,
ytick style={color=darkslategray38},
xtick={0, 20, 40, 60, 80, 100},
ytick={60, 65, 70, 75, 80, 85, 90, 95}
]
\addplot [semithick, purple, mark=*, mark size=1.2, mark options={solid,draw=white}]
table {%
6 84.6
9 71.4
12 73.5 
35 87.2
45 88
50 73
55 76.9
};
\addplot[color=red,mark=none,dashed] coordinates {
	(12,60)
	(12,95)
};
\node[] at (axis cs: 24,69) {{\tiny $|\mathcal{Y}^\ttarget|=12$}};
\addlegendentry{\autolabel}
\end{axis}

\end{tikzpicture}
\begin{tikzpicture}

\definecolor{darkslategray38}{RGB}{38,38,38}
\definecolor{lavender234234242}{RGB}{234,234,242}
\definecolor{lightgray204}{RGB}{204,204,204}
\definecolor{purple}{RGB}{128,0,128}

\begin{axis}[
axis background/.style={fill=lavender234234242},
axis line style={white},
legend cell align={left},
legend style={
  font=\tiny,
  fill opacity=0.8,
  draw opacity=1,
  text opacity=1,
  at={(0.97,0.80)},
  anchor=south east,
  draw=lightgray204,
  fill=white
},
title={{\tiny Epic Kitchens (D1 $\rightarrow$ D2)}},
height=0.53\linewidth,
width=0.53\linewidth,
tick align=outside,
x grid style={white},
xlabel=\textcolor{darkslategray38}{\(\displaystyle C\) (\# target clusters)},
xmajorgrids,
xmajorticks=true,
xtick pos=left,
xmin=20, xmax=160,
xtick style={color=darkslategray38},
y grid style={white},
ylabel=\textcolor{darkslategray38}{HOS (\%)},
ymajorgrids,
ymajorticks=true,
ytick pos=left,
ymin=15, ymax=55.1,
ytick style={color=darkslategray38},
xtick={30, 60, 90, 120, 150},
ytick={15, 25, 35, 45, 55}
]
\addplot [semithick, purple, mark=*, mark size=1.2, mark options={solid,draw=white}]
table {%
30 42.9
35 33
40 44.1
45 36.5
50 34.2
92 37
100 35.6
150 19.6
};
\addplot[color=red,mark=none,dashed] coordinates {
	(92,15)
	(92,55)
};
\node[] at (axis cs: 60,25) {{\tiny $|\mathcal{Y}^\ttarget|=92$}};
\addlegendentry{\autolabel}
\end{axis}

\end{tikzpicture}
    \vspace{-3mm}
    \caption{Impact of varying the \textbf{number of target clusters} $\mathcal{C}$.}
    \vspace{-3mm}
    \label{fig:abl-num-custers}
    \vspace{-3mm}
\end{figure}
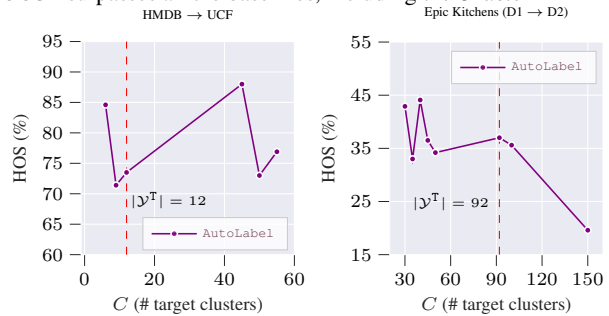

\vspace{1mm}

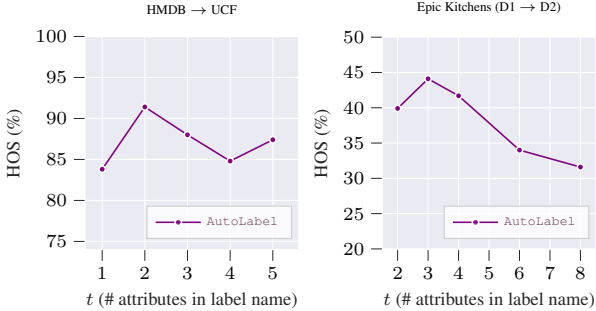
\begin{figure}[!t]
    \centering\scriptsize
\begin{tikzpicture}

\definecolor{darkslategray38}{RGB}{38,38,38}
\definecolor{lavender234234242}{RGB}{234,234,242}
\definecolor{lightgray204}{RGB}{204,204,204}
\definecolor{purple}{RGB}{128,0,128}

\begin{axis}[
axis background/.style={fill=lavender234234242},
axis line style={white},
legend cell align={left},
legend style={
  font=\tiny,
  fill opacity=0.8,
  draw opacity=1,
  text opacity=1,
  at={(0.97,0.05)},
  anchor=south east,
  draw=lightgray204,
  fill=white
},
title={{\tiny HMDB $\rightarrow$ UCF}},
height=0.53\linewidth,
width=0.53\linewidth,
tick align=outside,
x grid style={white},
xlabel=\textcolor{darkslategray38}{\(\displaystyle t\) (\# attributes in label name)},
xmajorgrids,
xmajorticks=true,
xtick pos=left,
xmin=0.6, xmax=5.6,
xtick style={color=darkslategray38},
y grid style={white},
ylabel=\textcolor{darkslategray38}{HOS (\%)},
ymajorgrids,
ymajorticks=true,
ytick pos=left,
ymin=74, ymax=100,
ytick style={color=darkslategray38},
xtick={0, 1, 2, 3, 4, 5},
ytick={75, 80, 85, 90, 95, 100}
]
\addplot [semithick, purple, mark=*, mark size=1.2, mark options={solid,draw=white}]
table {%
1 83.8
2 91.4
3 88
4 84.8
5 87.4
};
\addlegendentry{\autolabel}
\end{axis}

\end{tikzpicture}
\begin{tikzpicture}

\definecolor{darkslategray38}{RGB}{38,38,38}
\definecolor{lavender234234242}{RGB}{234,234,242}
\definecolor{lightgray204}{RGB}{204,204,204}
\definecolor{purple}{RGB}{128,0,128}

\begin{axis}[
axis background/.style={fill=lavender234234242},
axis line style={white},
legend cell align={left},
legend style={
  font=\tiny,
  fill opacity=0.8,
  draw opacity=1,
  text opacity=1,
  at={(0.97,0.05)},
  anchor=south east,
  draw=lightgray204,
  fill=white
},
title={{\tiny Epic Kitchens (D1 $\rightarrow$ D2)}},
height=0.53\linewidth,
width=0.53\linewidth,
tick align=outside,
x grid style={white},
xlabel=\textcolor{darkslategray38}{\(\displaystyle t\) (\# attributes in label name)},
xmajorgrids,
xmajorticks=true,
xtick pos=left,
xmin=1.5, xmax=8.5,
xtick style={color=darkslategray38},
y grid style={white},
ylabel=\textcolor{darkslategray38}{HOS (\%)},
ymajorgrids,
ymajorticks=true,
ytick pos=left,
ymin=19.9, ymax=50.1,
ytick style={color=darkslategray38},
xtick={1, 2, 3, 4, 5, 6, 7, 8},
ytick={20, 25, 30, 35, 40, 45, 50}
]
\addplot [semithick, purple, mark=*, mark size=1.2, mark options={solid,draw=white}]
table {%
2 39.9
3 44.1
4 41.7
6 34
8 31.6
};
\addlegendentry{\autolabel}
\end{axis}

\end{tikzpicture}
    \vspace{-3mm}
    \caption{Impact of varying the \textbf{number of attributes} $t$ \textbf{in candidate label name}.}
    \vspace{-5mm}
    \label{fig:abl-prompt-len}
\end{figure}

\noindent\textbf{Results on \textit{Epic-Kitchens}.} In~\cref{tab:3_kitchens} we report the results obtained on the more challenging \textit{Epic-Kitchens} (EK) benchmark. Due to the lack of space we only report the HOS scores for the six directions of EK. Detailed metrics can be found in the \supp. The observations on the EK benchmark is consistent with the trends observed in the \textit{HMDB$\leftrightarrow$UCF} benchmark. In particular, the \textbf{ActionCLIP} baseline achieves a much superior performance (+20.4 points) over the existing prior art \textbf{CEVT}. Moreover, when ActionCLIP is powered by our proposed \autolabel, the average HOS score further improves from 32.4 to 38.2. Note that the low overall performance on EK by every method hints at the fact that the actions depicted by the egocentric videos in EK are far more challenging. Despite the complexity in such actions, our proposed \autolabel sets a new state-of-the-art performance in \task, surpassing the competitors by a large margin.

Contrary to our previous observation, the \textbf{ActionCLIP}-\textit{Oracle} baseline is outperformed by our \autolabel. We hypothesize that the true target-private label names in the EK (\eg \texttt{flatten}) are quite generic and not descriptive enough to reject the target-private instances. Whereas, our \autolabel models the objects associated with such actions (\eg ``\texttt{rolling pin}'' for \texttt{flatten}), leading to an improved disambiguation of such action categories.

\subsection{Ablation Analysis}
\label{sec:ablation}

\noindent\textbf{Number of target clusters.} Given the significant role of the clustering step for semantically modeling the target domain, we show in~\cref{fig:abl-num-custers} the impact of varying the number of target clusters $\mathcal{C}$ on the HOS scores  for the \textit{HMDB$\rightarrow$UCF} and EK (D1 $\rightarrow$ D2) settings. The dashed red line indicates the ground truth number of target categories, \ie \(|\mathcal{C}| = |\mathcal{Y}^\ttarget|\). We can observe that, in both cases, the HOS peaks for values of \(|\mathcal{C}|\) in the range 35-45, which are higher and lower than the actual number of target classes for \textit{HMDB$\rightarrow$UCF} and \textit{Epic-Kitchens}, respectively. 

For \textit{HMDB$\rightarrow$UCF} these findings suggest that due to the wide diversity of actions, the discrimination of the framework is increased by modeling the target domain at a higher granularity. Whereas, for the EK, due to visual similarity of multiple actions, the discovered granularity in the label set is far less excessive than the ground truth action categories.



\vspace{1mm}

\noindent\textbf{Number of tokens in the candidate target-private labels.} In~\cref{fig:abl-prompt-len} we ablate on the number of attributes $t$ used to build the final target-private label candidate. In line with our hypothesis in~\cref{sec:attributes}, the ablation suggests that the candidate target labels composed by 2-3 attribute tokens to be a reasonable choice for effectively describing the candidate target action label names. In particular, when the value of $t$ increases significantly, the HOS scores for both the benchmarks show a steady decline due to the overcharacterization of the action labels with spurious attributes.

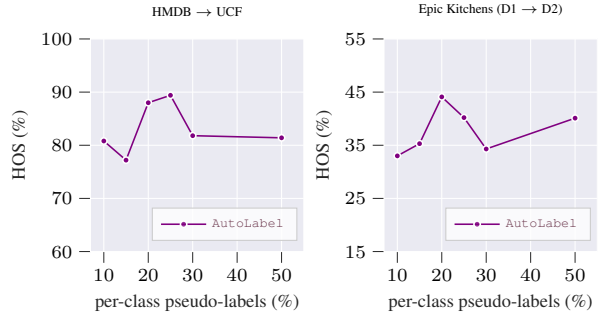
\begin{figure}[!h]
    \centering\scriptsize
\begin{tikzpicture}

\definecolor{darkslategray38}{RGB}{38,38,38}
\definecolor{lavender234234242}{RGB}{234,234,242}
\definecolor{lightgray204}{RGB}{204,204,204}
\definecolor{purple}{RGB}{128,0,128}

\begin{axis}[
axis background/.style={fill=lavender234234242},
axis line style={white},
legend cell align={left},
legend style={
  font=\tiny,
  fill opacity=0.8,
  draw opacity=1,
  text opacity=1,
  at={(0.97,0.05)},
  anchor=south east,
  draw=lightgray204,
  fill=white
},
title={{\tiny HMDB $\rightarrow$ UCF}},
height=0.53\linewidth,
width=0.53\linewidth,
tick align=outside,
x grid style={white},
xlabel=\textcolor{darkslategray38}{per-class pseudo-labels (\%)},
xmajorgrids,
xmajorticks=true,
xtick pos=left,
xmin=7, xmax=55,
xtick style={color=darkslategray38},
y grid style={white},
ylabel=\textcolor{darkslategray38}{HOS (\%)},
ymajorgrids,
ymajorticks=true,
ytick pos=left,
ymin=60, ymax=100.1,
ytick style={color=darkslategray38},
xtick={10, 20, 30, 40, 50},
ytick={60, 70, 80, 90, 100}
]
\addplot [semithick, purple, mark=*, mark size=1.2, mark options={solid,draw=white}]
table {%
10 80.8
15 77.2
20 88
25 89.4
30 81.8
50 81.4
};
\addlegendentry{\autolabel}
\end{axis}

\end{tikzpicture}
\begin{tikzpicture}

\definecolor{darkslategray38}{RGB}{38,38,38}
\definecolor{lavender234234242}{RGB}{234,234,242}
\definecolor{lightgray204}{RGB}{204,204,204}
\definecolor{purple}{RGB}{128,0,128}

\begin{axis}[
axis background/.style={fill=lavender234234242},
axis line style={white},
legend cell align={left},
legend style={
  font=\tiny,
  fill opacity=0.8,
  draw opacity=1,
  text opacity=1,
  at={(0.97,0.05)},
  anchor=south east,
  draw=lightgray204,
  fill=white
},
title={{\tiny Epic Kitchens (D1 $\rightarrow$ D2)}},
height=0.53\linewidth,
width=0.53\linewidth,
tick align=outside,
x grid style={white},
xlabel=\textcolor{darkslategray38}{per-class pseudo-labels (\%)},
xmajorgrids,
xmajorticks=true,
xtick pos=left,
xmin=7, xmax=55,
xtick style={color=darkslategray38},
y grid style={white},
ylabel=\textcolor{darkslategray38}{HOS (\%)},
ymajorgrids,
ymajorticks=true,
ytick pos=left,
ymin=15, ymax=55.1,
ytick style={color=darkslategray38},
xtick={10, 20, 30, 40, 50},
ytick={15, 25, 35, 45, 55}
]
\addplot [semithick, purple, mark=*, mark size=1.2, mark options={solid,draw=white}]
table {%
10 33
15 35.3
20 44.1
25 40.2
30 34.3
50 40.1
};
\addlegendentry{\autolabel}
\end{axis}

\end{tikzpicture}
    \vspace{-3mm}
    \caption{Impact of varying the \textbf{number of pseudo-labels}.}
    \vspace{-3mm}
    \label{fig:abl-num-pseudolabels}
    \vspace{-1mm}
\end{figure}

\vspace{1mm}

\noindent\textbf{Number of target pseudo-labels.} Finally, in~\cref{fig:abl-num-pseudolabels} we analyse the impact of the number of most confident pseudo-labels employed in order to fine-tune the model on the unlabelled target domain. In both the considered settings, we notice that the optimal percentage of pseudo-labels to be used for target fine-tuning lies around 20\%. Further addition of pseudo-labels leads to reduced performance, possibly due to the accumulation of noisy target pseudo-labels.

\vspace{-2mm}
\section*{Limitations}
\vspace{-1mm}

The proposed \autolabel models the target-private label name by indirectly modelling the objects and actors depicted in an action. Currently it can not disambiguate among different actions that involve the same objects and actors. For instance, ``\texttt{polo}'', ``\texttt{equestrian}'' and ``\texttt{skijoring}'' will all be described by the same candidate target-private label ``\texttt{horse person}''. Exploring the relationships between the actors and objects, along with their states, remains as a future work to overcome this limitation.
\vspace{-2mm}
\section{Conclusions}
\vspace{-1mm}

In this work we presented \autolabel, a CLIP-based automatic labelling framework for addressing the open-set unsupervised video domain adaptation. Our proposed \autolabel automatically discovers the candidate target-private class label names and extends the shared class names in order to equip CLIP with zero-shot prediction capabilities, which is indeed necessary for rejecting target-private instances. Empirically we show that \autolabel enabled CLIP models bode well for the \task task due to their rich representations and zero-shot detection capabilities.


\noindent \textbf{Acknowledgment.} This work was supported by the AI@TN project, the EU H2020 project SPRING funded by the European Commission under GA 87124, and by the Caritro Deep Learning lab of the ProM facility.

{\small
\bibliographystyle{ieee_fullname}
\bibliography{egbib}
}

\appendix

\twocolumn[\vspace*{2em}\centering\Large\bf%
{\Large Supplementary Material for} \\ {AutoLabel: CLIP-based framework for Open-set Video Domain Adaptation}%
\vspace*{4em}]

\setcounter{table}{0}
\renewcommand{\thetable}{A\arabic{table}}%
\setcounter{figure}{0}
\renewcommand{\thefigure}{A\arabic{figure}}%
\setcounter{equation}{0}
\renewcommand{\theequation}{A\arabic{equation}}%

The supplementary material is organized as follows: in Section \ref{sec:pseudo_labelling} we provide further details on our fine-tuning process on the target domain. In Section \ref{sec:attributes} we describe in depth the pipeline for attributes extraction and matching. In Section \ref{sec:pseudocode} we provide the pseudo-code for the most relevant routines of our \autolabel framework. In Section \ref{sec:datasets_statistics} we provide useful statistics of our considered benchmarks. In Section \ref{sec:baselines} we provide a more detailed description of the baseline methods included in our experimental evaluation. In Section \ref{sec:results} we report additional results and ablation study experiments.

\section{Target fine-tuning}
\label{sec:pseudo_labelling}

In this Section we provide details of how the target pseudo-labelling and consequent fine-tuning steps are carried out in our \autolabel framework. ActionCLIP \cite{wang2021actionclip} performs inference by projecting the test video to the CLIP space, and assigning the label corresponding to the textual prompts whose embedding is the most similar to the video embedding. In order to fine-tune on the target domain, we freeze the network and apply such inference step to the unlabelled target training batch, obtaining a pseudo-label for each target instance. After that, we filter out all those predictions that are not included in the top-\(k\%\) most confident ones for that specific label. In order to measure the confidence of a given pseudo-label, we consider the similarity in the CLIP space with the closest set of textual prompts. On the instances of the target training batch passing the filtering process, we simply carry out a standard supervised training step with the ActionCLIP loss. 

\section{Attributes extraction and matching}
\label{sec:attributes}

In this Section we detail the implementation of the attributes extraction and matching pipeline mentioned in the main paper. In particular, we provide the formal details of the \texttt{tfidf} and \(sim(\cdot,\cdot)\) functions from Sections 3.2.2 and 3.2.3, respectively. As mentioned in the paper, the off-the-shelf image captioning model ViLT \cite{kim2021vilt} extracts a set of attributes from a selection of frames in a given input video sequence. For our experiments, we set up the model in order to extract 5 attributes for each of 5 frames out of each video sequence. After that, we select the 5 most frequent attributes across the frame selection in order to build the final set of attributes for the sequence.


\subsection{\texttt{tfidf} function}
\label{subsec:tfidf}

After the extraction carried out by ViLT, we are provided with a set of attributes for each video sequence. The following steps demand the extraction of a set of attributes for a given class (source domain) and for a given video cluster (target domain). In both cases the pipeline is the same, and we only change the set of instances given as input. Given the sets of instances, we apply the \texttt{tfidf} module, which is implemented as follows. We firstly compute the most frequent attributes across all the input instances; at this point, we compute the \textit{Term-frequency and Inverse Document Frequency (tf-idf)} \cite{FRIEDMAN2013765, MASUDA2011281} score of each attribute. Given a set of text documents and the corresponding token vocabulary, the \textit{tf-idf} value of a given token with respect of a given document is designed in order to quantify how relevant that token is for that document. Formally, this score is defined as the product of two different statistics, namely \textit{Term frequency (tf)} and \textit{Inverse document frequency (idf)}, defined as follows:

\begin{equation}
    tf(t,d) = \frac{f_{t,d}}{\sum_{t' \in d} f_{t', d}}
\end{equation}

\begin{equation}
    idf(t,D) = \log \frac{N}{|\{d \in D : t \in d\}}
\end{equation}

\begin{equation}
    tfidf(t,d,D) = tf(t,d) \cdot idf(t,D)
\end{equation}
where \(f_{t,d}\) is the raw count of of term \(t\) in the document \(d\), i.e., the number of occurrences of term \(t\) in \(d\). \(N\) is the total number of documents considered, and \(D\) is the set of documents. 

The conceptual intuition behind this score lies in the fact that a given term is most likely to be relevant for a given document if (i) it occurs often in the document and (ii) it occurs seldom in any other document. In our case, we consider one document for each class (\(N = K\), source domain) or for each cluster (\(N = C\), target domain): each document comprises the most common attributes across that specific set of instances. Consequently, by thresholding the \textit{tf-idf} of each attribute, we end up with the most relevant terms for each source class and for each target cluster. We set this threshold to $0.5$ for all experiments.

\subsection{Matching}
\label{sec:matching}

Once provided with a set of relevant terms for each source class and for each target cluster, we carry out the matching step as described in the main document. This step relies on the \(sim(\cdot,\cdot)\) function, which takes as input two sets of attributes, ordered by confidence during the extraction, and computes a similarity score. Given \(a_s\) attributes in the source set 
, this function defines \(a_s\) weights in decreasing order normalized between 0 and 1. For each common occurrences, it adds up to the score value (initially 0) the weight corresponding the absolute distance between the positions of such occurrences in the two input sets. Informally, the intuition behind this score consists in the idea of taking into account both the number of co-occurrences of attributes between the two sets and their position, the latter accounting for their frequency in the corresponding input set. At the end of this loop, the score is again normalized and used to fill the \(S\) matrix mentioned in Sec. 3.2.3.

\section{Pseudo-code}
\label{sec:pseudocode}

In this Section we provide the pseudo-code for the different modules of our framework. In particular, Alg. \ref{alg:attr_extraction} presents the attribute extraction step presented in Section 3.2.1 of the main document; Alg. \ref{alg:class_discovery} presents the discovery process of candidate target classes presented in 3.2.2, Alg. \ref{alg:sim_function} details the similarity function \(sim(\cdot,\cdot)\) referenced in 3.2.3 and Alg. \ref{alg:attr_matching} provides the pipeline for the attribute matching process described in 3.2.3.


\begin{algorithm}
\caption{Attribute extraction}\label{alg:attr_extraction}
    \KwInput{\text{Source video sequences} $\bfX^\tsource$, \text{Target video sequences} $\bfX^\ttarget$, \text{Prompt} $z$, \text{ViLT model} \texttt{ViLT()}, \text{Number of selected frames} $F$, Number of selected attributes $k$, \text{\textit{tf-idf} module} \texttt{tfidf}}
    \KwOutput{\text{Set of source attributes} $\bar{\Lambda}^{l^\tsource}$, \text{Set of target attributes} $\bar{\Lambda}^{\ttarget}$}
    \BlankLine
    \For{$\bfX \in \{\bfX^\tsource, \bfX^\ttarget\}, d \in \{l^\tsource, \ttarget\}$}{
        \For{$i \gets 0$ \KwTo $|\bfX|$}{
            $\bfx \gets \bfX[i]$ \\
            $\bf{video\_attributes} \gets [\ ]$ \\
            \For{$j \in F$}{
                $\mathcal{A}(\bfx_j) \gets$ \texttt{ViLT}\((\bfx_j, z)\) \\
                \text{Append attributes in} $\mathcal{A}(\bfx_j)$ \text{to} $\bf{video\_attributes}$
            }
        }
        $\bf{mc}$ \(\gets \text{argtop}_k(\bf{video\_attributes})\) \\
        $\bf{filtered} \gets$ \texttt{tfidf}\((\bf{mc})\) \\
        \text{Add attributes in} $\bf{filtered}$ \text{to} $\bar{\Lambda}^d$
        
    }
\end{algorithm}

\begin{algorithm}
\caption{Discovering candidate classes}\label{alg:class_discovery}
    \KwInput{\text{Target video sequences} $\bfX^\ttarget$, \text{Video encoder} $G_V$, Number of target clusters $|\mathcal{C}|$, \text{Set of target attributes} $\bar{\Lambda}^{\ttarget}$, Clustering function $Cluster$}
    \KwOutput{\text{Target candidate labels} $\mathcal{Y}^{\tcand}$}
    \BlankLine
    $\bfv^\ttarget$ \(\gets G_V(\bfX^\ttarget)\) \\
    $\mathcal{C}$ \(\gets Cluster(\bfv^\ttarget)\) \\
    $\mathcal{Y}^{\tcand}$ \(\gets \O\) \\
    \For{$c \gets 0$ \KwTo $|\mathcal{C}|$}{
        $\bar{\Lambda}^{c, \ttarget} \gets$ \text{Attributes for videos belonging to cluster} $c$ \\
        \(l^{\tcand}_c = \bar{\Lambda}^{c, \ttarget}_1 || \dots || \bar{\Lambda}^{c, \ttarget}_t \) \\
        \text{Add} \(l^{\tcand}_c\) \text{to} $\mathcal{Y}^{\tcand}$ \\
    }
    \BlankLine
    
\end{algorithm}

\begin{algorithm}
\caption{Similarity function $sim(\cdot,\cdot)$}\label{alg:sim_function}
    \KwInput{\text{Set of source attributes} $\bar{\Lambda}^{l^\tsource}$, \text{Set of target attributes} $\bar{\Lambda}^{\ttarget}$}
    \KwOutput{\text{Similarity score $s$}}
    \BlankLine
    \Comment{Compute normalized weights }
    $\bf{ref}$ \(\gets reverse(range(len(\bar{\Lambda}^{l^\tsource_i})))\) \\
    $\bfw$ \(\gets\) ($\bf{ref}$ \(-\) \(min(\)$\bf{ref}$\()) / (max(\)\bf{ref}\() - min(\bf{ref}))\) \\
    \BlankLine
    \Comment{Incrementally compute score}
    $s$ \(\gets 0\) \\
    \For{$i_s \gets 0$ \KwTo $len(\bar{\Lambda}^{l^\tsource})$}{
        \For{$i_t \gets 0$ \KwTo $len(\bar{\Lambda}^{\ttarget})$}{
            \If{\(\bar{\Lambda}^{l^\tsource}[i_s] = \bar{\Lambda}^{\ttarget}[i_t]\)}{
                $s$ \(\gets\) $s$ \(+\) $\bfw$\([abs(i_t-i_s)]\) \\
            }
        }
    }
    $s$ \(\gets s/len(\bar{\Lambda}^{l^\tsource_i})\) \\
    
\end{algorithm}

\begin{algorithm}
\caption{Attribute matching}\label{alg:attr_matching}
    \KwInput{\text{Target candidate labels} $\mathcal{Y}^{\tcand}$, \text{Similarity function} $sim(\cdot,\cdot)$, \text{Threshold} $\gamma$, \text{Number of shared classes} $K$, \text{Number of target clusters} $|\mathcal{C}|$, \text{Number of tokens} $t$}
    \KwOutput{\text{Target private labels} $\mathcal{Y}^{\tpriv}$}
    \BlankLine
    $\mathcal{Y}^{\tpriv}$ \(\gets \O\) \\
    \For{$i \gets 0$ \KwTo $|\mathcal{C}|$}{
        \(\bf{match} \gets \bf{False}\) \\
        \For{$j \gets 0$ \KwTo $K$}{
            \If{\(sim(\)$\bar{\Lambda}^{l^\tsource_j}$, $\bar{\Lambda}^{i, \ttarget}$\() < \gamma\)}{
            \(\bf{match} \gets \bf{True}\)
            }
        }
        \If{\(\neg\bf{match}\)}{
            \(l^{\tcand} = \mathcal{Y}^{\tpriv}[i]\) \\
            \text{Add} \(l^{\tcand}\) \text{to} $\mathcal{Y}^{\tpriv}$ \\
        }
    }
\end{algorithm}

\section{Datasets statistics}
\label{sec:datasets_statistics}

In this Section we provide detailed statistics about the benchmarks considered in our experimental evaluation. In particular, we report for each dataset in Table \ref{tab:datasets_statistics} the number of shared and private classes and the number of source training, target training and test samples.

\begin{table*}[]
    \centering
    \small
    \begin{tabular}{l|c|c|c|c|c}
        \toprule
        \textbf{Dataset} & \# \textbf{shared classes} & \# \textbf{private classes} & \# \textbf{source train samples} & \# \textbf{target train samples} & \# \textbf{test samples} \\
        \midrule
        \textit{HMDB} & 6 & 6 & 375 & 781 & 337 \\
        \textit{UCF} & 6 & 6 & 865 & 1438 & 571 \\
        \textit{EK-D1} & 8 & 75 & 1543 & 2021 & 625 \\
        \textit{EK-D2} & 8 & 84 & 2495 & 3755 & 885 \\
        \textit{EK-D3} & 8 & 82 & 3897 & 5847 & 1230 \\
        \bottomrule
    \end{tabular}
    \caption{Statistics of the considered benchmarks for the experimental evaluation}
    \label{tab:datasets_statistics}
\end{table*}

\section{Baseline details}
\label{sec:baselines}


In this Section we provide additional details about the baseline methods we implemented autonomously. 

\paragraph{CEVT-CLIP \cite{chen2021conditional}} 
This baseline has been implemented by simply modifying the original code provided by the authors in~\cite{chen2021conditional}, replacing the ResNet \cite{resnet} backbone with the ActionCLIP \cite{wang2021actionclip} encoder. 

\paragraph{ActionCLIP \cite{wang2021actionclip}} 
This baseline is obtained by modifying our own framework, itself based on the ActionCLIP architecture, in order to apply a different open-set rejection protocol. In order to make the choice of whether to assign a known or unknown label to a test target sample, this method simply thresholds the similarity, in the CLIP space, between the video embedding and the closest set of label prompts. This threshold has been set to 0.9 for \textit{HMDB$\leftrightarrow$UCF} and to 0.5 for \textit{Epic-Kitchens}. 

\paragraph{ActionCLIP-ZOC \cite{esmaeilpour2022zero}} 
This baseline is implemented as a modification of the open-set rejection protocol of our method: instead of extending the target label set with newly discovered labels extracted by unmatched target clusters, this method extends it for each individual test instance with the names of the objects detected by ViLT \cite{kim2021vilt} in that specific sequence. The detection process is carried out in the same way as in \autolabel. 

\paragraph{ActionCLIP-\textit{Oracle} \cite{fort2021exploring}} 
We implement this baseline by extending the label set with the ground truth names of the target private categories. For fair comparison, all hyperparameters for these baselines match those employed for \autolabel in each setting.

\section{Additional results}
\label{sec:results}

\subsection{Detailed \textit{Epic-Kitchens} results}
\label{subsec:ek_results}

We report in Table \ref{tab:ek_results} the complete results of our method and its competitors on the \textit{Epic-Kitchens} setting, including the \textbf{ALL}, $\textbf{OS}^*$ and \textbf{UNK} metrics, omitted in the main document for space issues. it is possible to observe in the complete Table that, especially when compared to the \textit{HMDB$\leftrightarrow$UCF} case, this benchmark is characterized by a significant instability. In particular, it is evident that, across different methods considered, the \textbf{HOS} score is affected by a strong tendency of most methods to either over-accept, resulting in a higher $\textbf{OS}^*$ score, or over-reject, producing a higher \textbf{UNK} score. However, it is possible to observe that the results obtained with our proposed \autolabel method, when compared to most competitors, are characterized by a better balance between $\textbf{OS}^*$ and \textbf{UNK}, indicating a more controlled training process. 

\begin{table*}[h!]
    \centering
    \begin{tabular}{|l|ccc|c|ccc|c|ccc|c|}
        \toprule
        \textbf{Setting} $\rightarrow$ & \multicolumn{4}{|c|}{D2$\rightarrow$D1} & \multicolumn{4}{|c|}{D3$\rightarrow$D1} & \multicolumn{4}{|c|}{D1$\rightarrow$D2} \\
        \midrule
        \textbf{Method} $\downarrow$ & \textbf{ALL} & $\textbf{OS}^*$ & \textbf{UNK} & \textbf{HOS} & \textbf{ALL} & $\textbf{OS}^*$ & \textbf{UNK} & \textbf{HOS} & \textbf{ALL} & $\textbf{OS}^*$ & \textbf{UNK} & \textbf{HOS} \\
        \midrule
        CEVT \cite{chen2021conditional} & \textbf{30.5} & 7.2 & \textbf{76.8} & 13.2 & \textbf{31.8} & 8.1 & \textbf{76.8} & 14.7 & 18.7 & 4.5 & 67.4 & 8.4 \\
        CEVT-CLIP \cite{chen2021conditional} & 26.8 & 7.3 & 68.9 & 13.2 & 24.4 & 10.0 & 67.8 & 17.3 & 16.6 & 7.3 & 71.8 & 13.3 \\
        \actionclipbaseline \cite{wang2021actionclip} & 24.6 & \textbf{32.2} & 48.1 & 31.3 & 19.5 & 29.2 & 27.5 & 28.3 & 21.3 & 25.6 & \textbf{74.5} & 38.1 \\
        ActionCLIP-\aaai\cite{esmaeilpour2022zero} & 22.0 & 18.4 & 43.6 & 25.9 & 20.9 & 29.2 & 24.7 & 26.8 & \textbf{23.6} & 24.7 & 44.4 & 31.7 \\
        \autolabel (ours) & 28.5 & 26.1 & 52.3 & \textbf{34.8} & 29.6 & \textbf{30.0} & 52.9 & \textbf{38.3} & 23.3 & \textbf{33.9} & 63.1 & \textbf{44.1} \\
        \midrule
        \neurips \cite{fort2021exploring} & \textit{25.6} & \textit{23.8} & \textit{55.0} & \textit{33.2} & \textit{21.9} & \textit{26.0} & \textit{45.5} & \textit{33.1} & \textit{31.7} & \textit{33.1} & \textit{42.1} & \textit{37.1} \\
        \midrule
        \textbf{Setting} $\rightarrow$ & \multicolumn{4}{|c|}{D3$\rightarrow$D2} & \multicolumn{4}{|c|}{D1$\rightarrow$D3} & \multicolumn{4}{|c|}{D2$\rightarrow$D3} \\
        \midrule
        \textbf{Method} $\downarrow$ & \textbf{ALL} & $\textbf{OS}^*$ & \textbf{UNK} & \textbf{HOS} & \textbf{ALL} & $\textbf{OS}^*$ & \textbf{UNK} & \textbf{HOS} & \textbf{ALL} & $\textbf{OS}^*$ & \textbf{UNK} & \textbf{HOS} \\
        \midrule
        CEVT \cite{chen2021conditional} & 25.2 & 8.9 & \textbf{78.5} & 16.0 & 21.6 & 4.3 & \textbf{71.0} & 8.1 & 25.5 & 6.1 & \textbf{77.7} & 11.3 \\
        CEVT-CLIP \cite{chen2021conditional} & 21.3 & 8.0 & 67.4 & 14.3 & 21.5 & 5.5 & 69.1 & 10.2 & 19.8 & 5.5 & 65.2 & 10.1 \\
        \actionclipbaseline \cite{wang2021actionclip} & 24.6 & 35.8 & 55.2 & 43.4 & 26.3 & 20.4 & 50.4 & 29.0 & 30.6 & 16.7 & 44.2 & 24.2 \\
        ActionCLIP-\aaai\cite{esmaeilpour2022zero} & 23.8 & 34.1 & 52.5 & 41.3 & 23.5 & 21.3 & 41.0 & 28.0 & \textbf{31.1} & \textbf{24.1} & 22.2 & 23.1 \\
        \autolabel (ours) & 25.7 & \textbf{39.9} & 68.4 & \textbf{50.4} & \textbf{29.6} & \textbf{28.5} & 36.2 & \textbf{31.9} & 27.7 & 21.1 & 50.8 & \textbf{29.8} \\
        \midrule
        \neurips \cite{fort2021exploring} & \textit{16.3} & \textit{31.7} & \textit{75.4} & \textit{44.6} & \textit{21.2} & \textit{17.8} & \textit{37.6} & \textit{24.2} & \textit{28.4} & \textit{18.8} & \textit{62.8} & \textit{28.9} \\
        \bottomrule
    \end{tabular}
    \caption{Results of all considered methods for the \textit{Epic-Kitchens} settings. We include in this Table all the open-set metrics, included those omitted from the main document for space issues. Our proposed \autolabel method is shown to achieve the best \textbf{HOS} score in all settings by achieving an effective balance of $\textbf{OS}^*$ and \textbf{UNK}.}
    \label{tab:ek_results}
\end{table*}

\subsection{Ablation analysis}
\label{subsec:ablation}

\begin{figure}[!t]
    \centering
    \includegraphics[width=0.6\columnwidth]{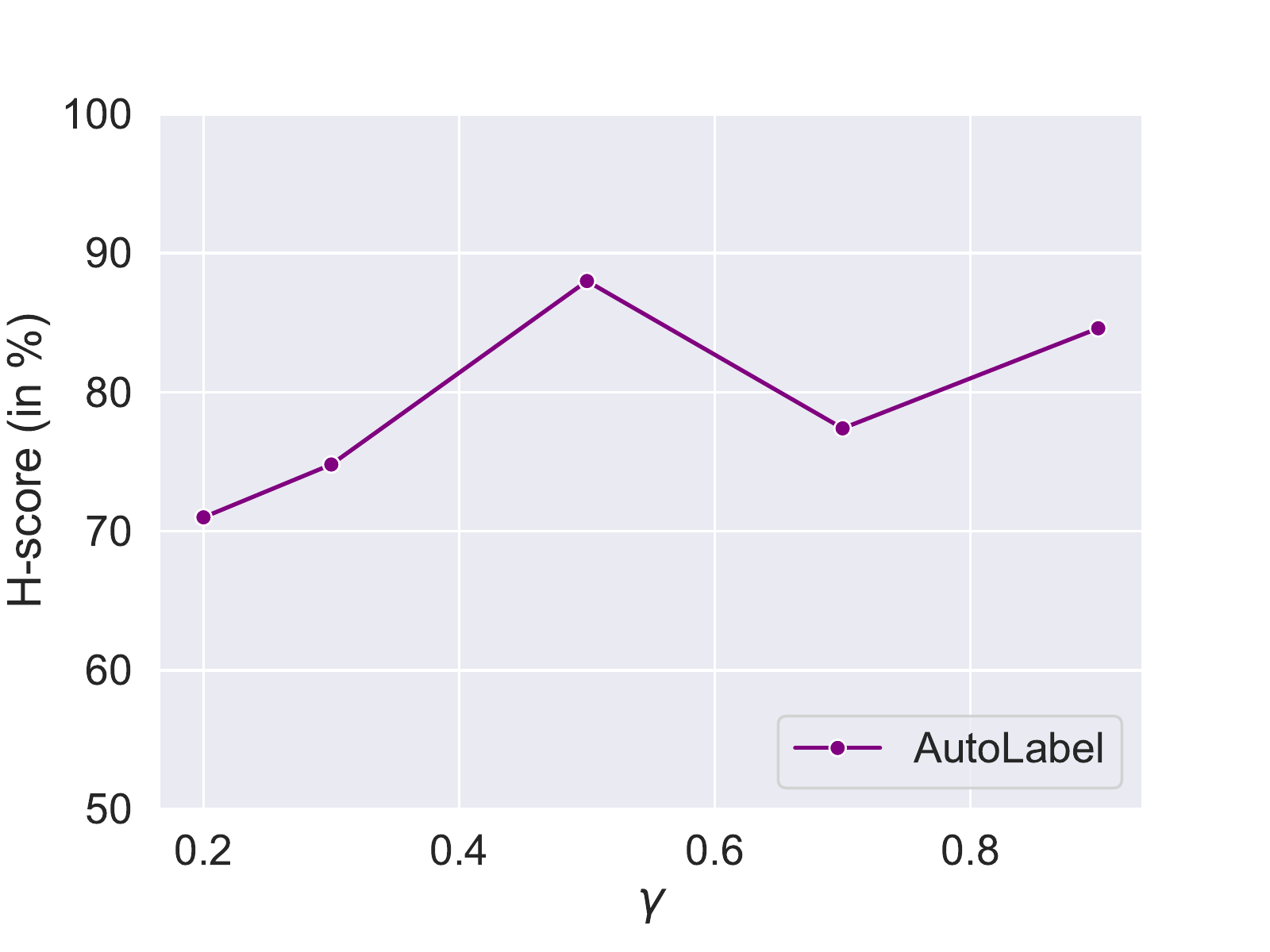}
    \caption{Sensitivity study on the threshold $gamma$ for \textit{HMDB$\rightarrow$UCF}} 
    \label{fig:abl-gamma-hu}
\end{figure}

\begin{figure}[!t]
    \centering
    \includegraphics[width=0.6\columnwidth]{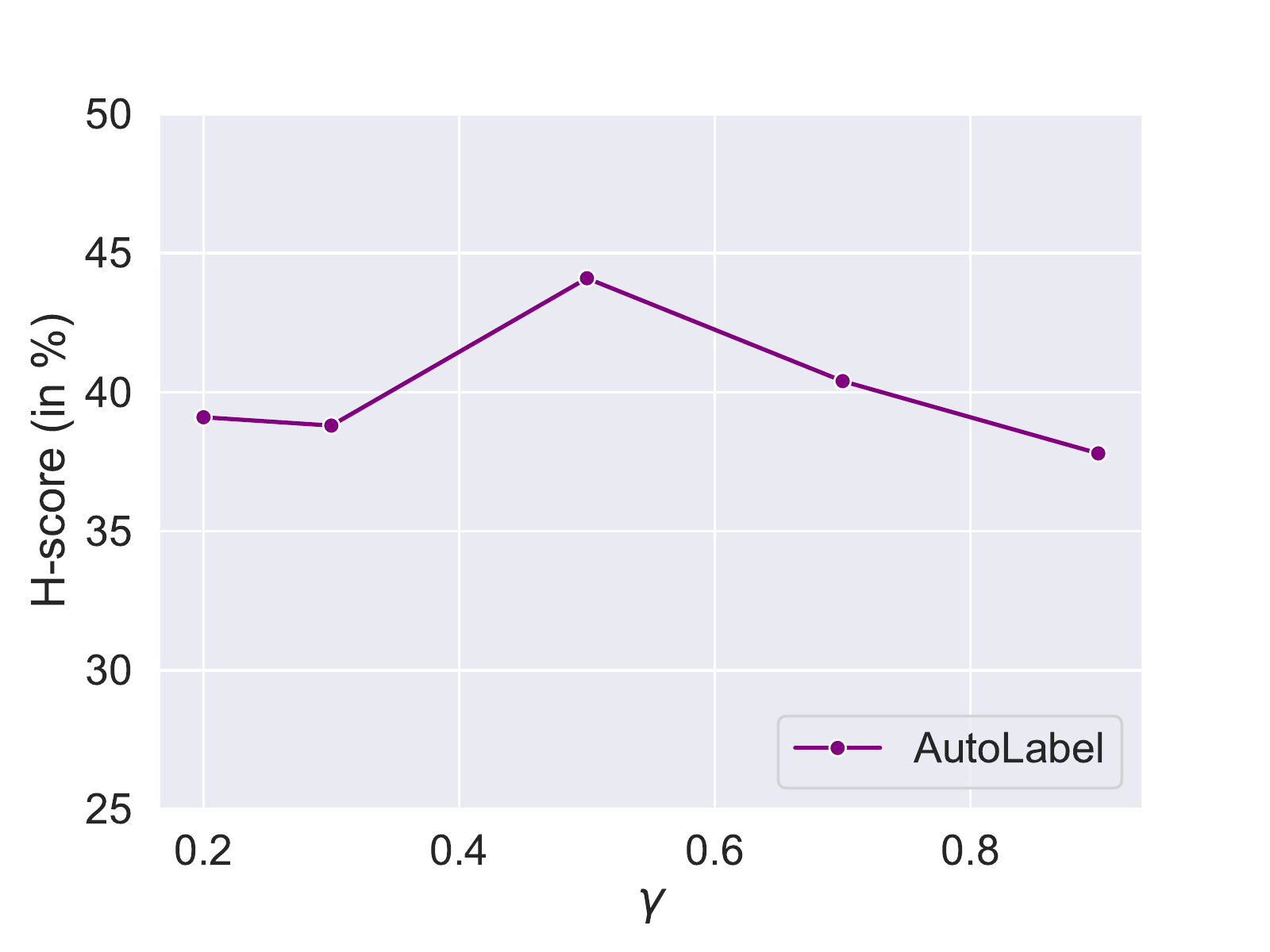}
    \caption{Sensitivity study on the threshold $gamma$ for \textit{Epic-Kitchens D1$\rightarrow$D2}} 
    \label{fig:abl-gamma-ek}
\end{figure}

We provide in this Section a further ablation analysis omitted from the main document for space issue. In particular, we report a sensitivity score on the matching threshold $\gamma$ with respect to the reference \textbf{HOS} metric, for \textit{HMDB$\rightarrow$UCF} (Fig. \ref{fig:abl-gamma-hu}) and for \textit{Epic-Kitchens D1$\rightarrow$D2} (Fig. \ref{fig:abl-gamma-ek}). From this study, it emerges that the score consistently oscillates around 80\% for \textit{HMDB$\rightarrow$UCF} and around 40\% for \textit{Epic-Kitchens}.

\subsection{Discovered candidate classes}
\label{subsec:class_discovery}


We provide in this Section an overview of the ground-truth and discovered target-private classes for the \textit{HMDB$\rightarrow$UCF} and \textit{Epic-Kitchens D1$\rightarrow$D2} settings, in Tables \ref{tab:discovered_hu} and \ref{tab:discovered_ek}, respectively. In the left column of the Tables we report the actual names of the private classes of the target domain, and on the right one we report the names of the candidate target-private labels identified by our proposed \autolabel framework, which are composed by concatenating the most relevant attributes extracted from each cluster that was not matched with any shared class. We can firstly observe that the discovered classes on \textit{HMDB$\rightarrow$UCF} show a significant diversity, especially when looking at the first attributes for each candidate label name, which are the most relevant ones. On the other hand, discovered classes on \textit{Epic-Kitchens} appear to be significantly more noisy and generic. As mentioned in the main document, we associate this behavior to the fact that video sequences in each domain of the \textit{Epic-Kitchens} dataset are all constrained to the same kitchen environment, thus characterized by the same (or similar) objects across multiple categories.

\begin{table}[h!]
    \centering
    \begin{tabular}{|c|c|}
        \toprule
        \textbf{Ground truth} & \textbf{Discovered} \\
        \midrule
        pushup & water AND horse AND fence \\
        ride bike & rope AND table AND table AND window \\
        ride horse & bike AND street AND car \\
        shoot ball & basketball AND building AND fence \\
        shoot bow & rock AND rope AND window \\
        walk & road AND bike AND car \\
        & sign AND net AND court \\
        & horse AND field AND building \\
        & floor AND chair AND table \\
        & refrigerator AND bed AND door \\
        & field AND dog AND grass \\
        & boxers AND men AND referee \\
        & horse AND building AND fence \\
        & dog AND grass AND fence \\
        & hoop AND basketball AND net \\
        & rack AND door AND mirror \\
        & house AND grass AND building \\
        & soccer AND field AND net \\
        & stick AND grass AND fence \\
        \bottomrule
    \end{tabular}
    \caption{List of the actual names of the ground truth target private classes (left) and a selection of candidate target-private label names identified by \autolabel (right) for the \textit{HMDB$\rightarrow$UCF} setting}
    \label{tab:discovered_hu}
\end{table}

\subsection{Cluster attributes}
\label{sec:cluster_attrs}

We show in Tables \ref{tab:attributes_hu_1} and Tables \ref{tab:attributes_hu_2}, respectively, two examples of the attributes extracted from sample target cluster for the \textit{HMDB$\rightarrow$UCF} setting, along with the final target description obtained with the \texttt{tfidf} module. It is possible to observe in these tables how the final attributes are able to reduce redundancy and provide an effective description for the candidate unknown class.

\begin{table}[h!]
    \centering
    \begin{tabular}{|c|c|c|}
        \toprule
        \multicolumn{2}{|c|}{\textbf{Original cluster attributes}} & \textbf{Final cluster attributes} \\
        \midrule
        \textcolor{blue}{horse} & fence & \textcolor{blue}{horse}  \\
        \textcolor{red}{fence} & \textcolor{green}{person} & \textcolor{green}{person} \\
        field & people & \textcolor{red}{fence} \\
        man & dirt & \textcolor{orange}{man} \\
        grass & \textcolor{blue}{horse} & \textcolor{purple}{tree} \\
        \textcolor{blue}{horse} & sand & \textcolor{gray}{water} \\
        zebra & \textcolor{blue}{horse} & \\
        mountain & \textcolor{green}{person} & \\
        bush & \textcolor{red}{fence} & \\
        \textcolor{blue}{horse} & \textcolor{orange}{man} & \\
        \textcolor{purple}{tree} & sand & \\
        \textcolor{gray}{water} & \textcolor{blue}{horse} & \\
        \textcolor{red}{fence} & beach & \\
        \textcolor{green}{person} & people & \\
        \textcolor{gray}{water} & \textcolor{orange}{man} & \\
        \textcolor{green}{person} & hat & \\
        \textcolor{blue}{horse} & shirt & \\
        bush & sky & \\
        \textcolor{green}{person} & \textcolor{blue}{horse} & \\
        \textcolor{orange}{man} & mountain & \\
        road & bush & \\
        \textcolor{blue}{horse} & sand & \\
        zebra & \textcolor{green}{person} & \\
        \textcolor{blue}{horse} & \textcolor{gray}{water} & \\
        beach & \textcolor{orange}{man} & \\
        \textcolor{gray}{water} & \textcolor{red}{fence} & \\
        woman & \textcolor{blue}{horse} & \\
        building & \textcolor{purple}{tree} & \\
        \textcolor{blue}{horse} & \textcolor{green}{person} & \\
        beach & bunch & \\
        \textcolor{blue}{horse} & \textcolor{purple}{tree} & \\
        sand & \textcolor{red}{fence} & \\
        hat & \textcolor{green}{person} & \\
        water & car & \\
        beach & \textcolor{blue}{horse} & \\
        
        \bottomrule
    \end{tabular}
    \caption{List of original and final attributes extracted from a sample target cluster for the \textit{HMDB$\rightarrow$UCF} setting. We emphasize each of the final attributes in a different color in order to highlight occurrences among the original ones}
    \label{tab:attributes_hu_1}
\end{table}

\begin{table*}[h!]
    \centering
    \begin{tabular}{|c|c|c|c|}
        \toprule
        \multicolumn{3}{|c|}{\textbf{Original cluster attributes}} & \textbf{Final cluster attributes} \\
        \midrule
        \textcolor{blue}{ball} & \textcolor{blue}{ball} & net & \textcolor{blue}{ball} \\
        \textcolor{red}{hoop} & people & \textcolor{orange}{basketball} & \textcolor{green}{male} \\
        \textcolor{green}{male} & sign & \textcolor{blue}{ball} & \textcolor{orange}{basketball}\\
        \textcolor{orange}{basketball} & light & gym & \textcolor{red}{hoop} \\
        white & kite & \textcolor{green}{male} & \textcolor{purple}{net} \\
        \textcolor{blue}{ball} & \textcolor{green}{male} & door & \textcolor{gray}{court} \\
        \textcolor{green}{male} & female & rack & \\
        \textcolor{purple}{net} & \textcolor{blue}{ball} & \textcolor{blue}{ball} & \\
        \textcolor{gray}{court} & rack & female & \\
        \textcolor{red}{hoop} & boy & \textcolor{green}{male} & \\
        gym & \textcolor{blue}{ball} & people & \\
        \textcolor{red}{hoop} & \textcolor{orange}{basketball} & \textcolor{orange}{basketball} & \\
        \textcolor{blue}{ball} & men & \textcolor{red}{hoop} & \\
        \textcolor{green}{male} & \textcolor{gray}{court} & \textcolor{blue}{ball} & \\
        \textcolor{orange}{basketball} & \textcolor{purple}{net} & \textcolor{green}{male} & \\
        \textcolor{blue}{ball} & \textcolor{blue}{ball} & door & \\
        \textcolor{green}{male} & \textcolor{red}{hoop} & \textcolor{red}{hoop} & \\
        rack & \textcolor{green}{male} & \textcolor{blue}{ball} & \\
        people & \textcolor{orange}{basketball} & \textcolor{green}{male} & \\
        table & white & \textcolor{green}{male} & \\
        \textcolor{blue}{ball} & \textcolor{blue}{ball} & \textcolor{gray}{court} & \\
        \textcolor{red}{hoop} & \textcolor{purple}{net} & man & \\
        \textcolor{purple}{net} & people & \textcolor{red}{hoop} & \\
        basket & \textcolor{gray}{court} & \textcolor{purple}{net} & \\
        person & \textcolor{orange}{basketball} & \textcolor{orange}{basketball} & \\
        \textcolor{red}{hoop} & \textcolor{orange}{basketball} & \textcolor{gray}{court} & \\
        \textcolor{blue}{ball} & \textcolor{blue}{ball} & \textcolor{gray}{court} & \\
        \textcolor{green}{male} & men & men & \\
        gym & \textcolor{purple}{net} & \textcolor{green}{male} & \\
        female & basket & \textcolor{red}{hoop} & \\
        \textcolor{blue}{ball} & \textcolor{blue}{ball} & basket & \\
        \textcolor{blue}{ball} & \textcolor{orange}{basketball} & gym & \\
        \bottomrule
    \end{tabular}
    \caption{List of original and final attributes extracted from a sample target cluster for the \textit{HMDB$\rightarrow$UCF} setting. We emphasize each of the final attributes in a different color in order to highlight occurrences among the original ones}
    \label{tab:attributes_hu_2}
\end{table*}

\subsection{Output visualization}
\label{sec:output_visualization}

\begin{figure}[!t]
    \centering
    \includegraphics[width=\columnwidth]{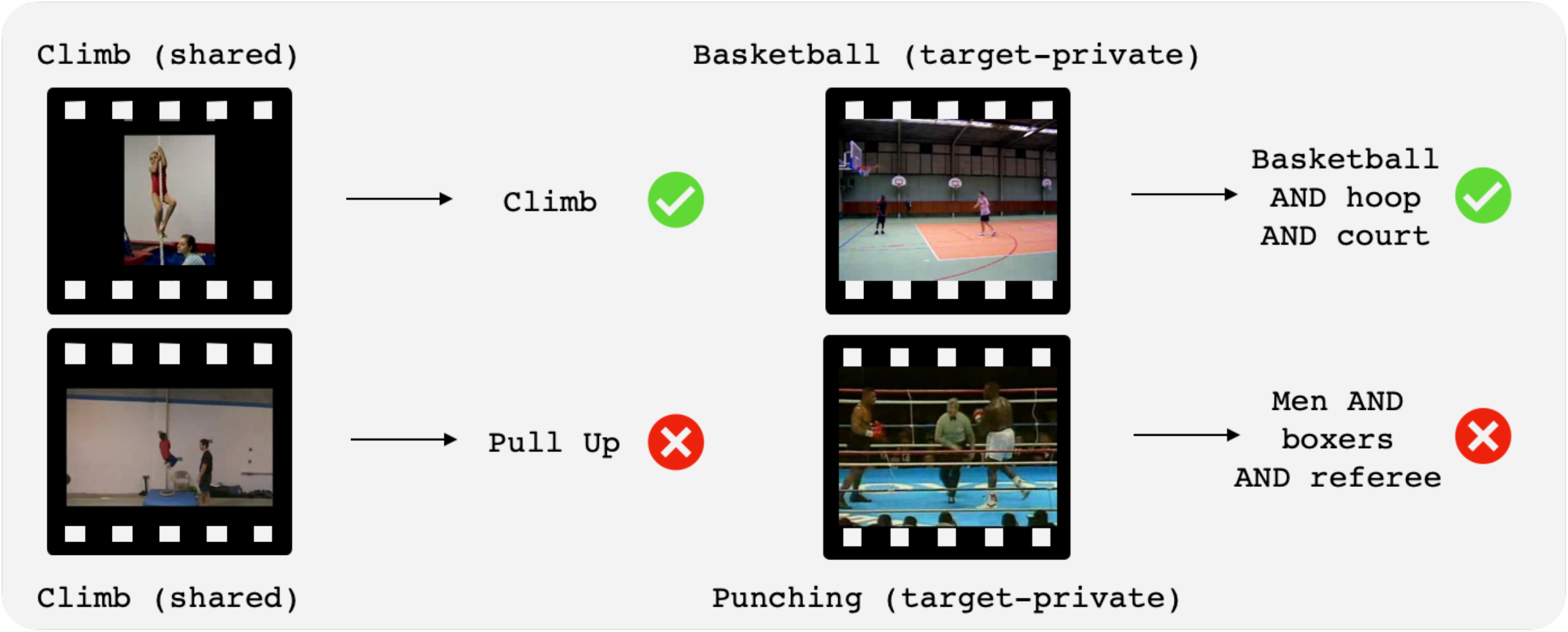}
    \caption{Example of correct and incorrect predictions of \autolabel on both shared and private categories on the \textit{HMDB$\rightarrow$UCF} setting}
    \label{fig:output_visualization_hu}
\end{figure}

\begin{figure}[!t]
    \centering
    \includegraphics[width=\columnwidth]{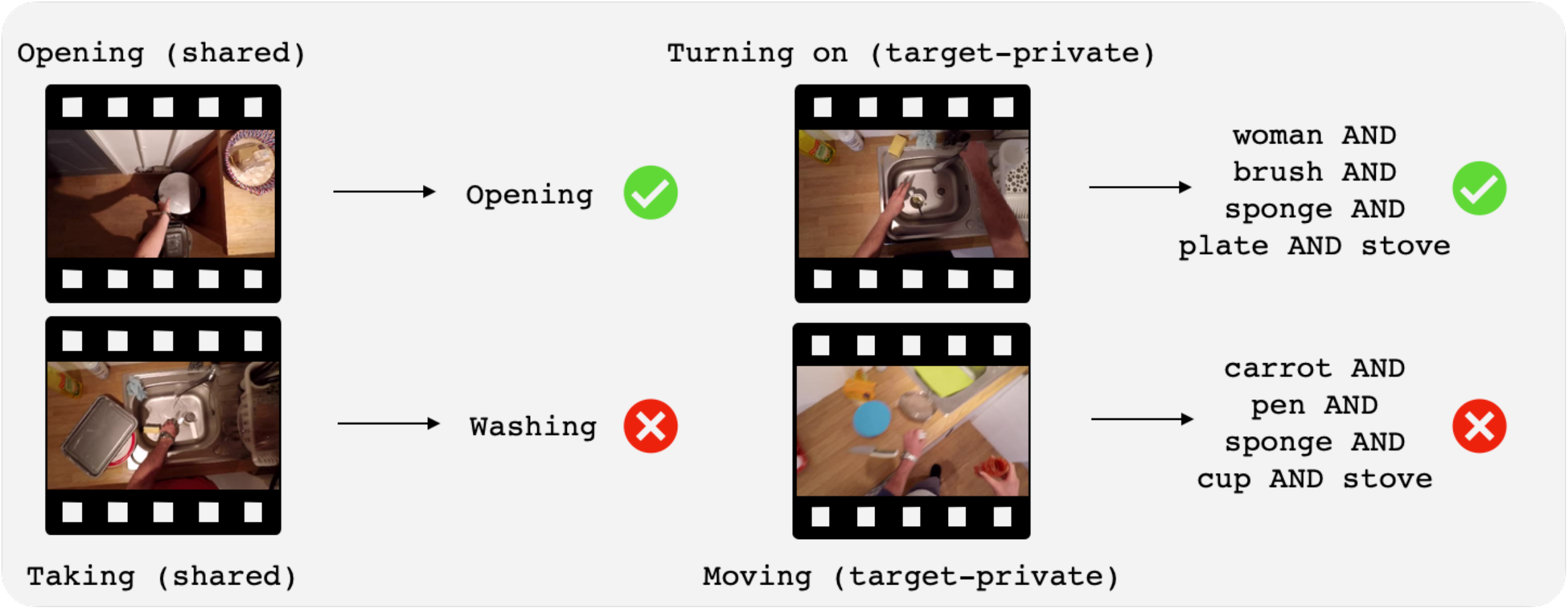}
    \caption{Example of correct and incorrect predictions of \autolabel on both shared and private categories on the \textit{Epic-Kitchens D1$\rightarrow$D2} setting} 
    \label{fig:output_visualization_ek}
\end{figure}

We provide in this Section examples of correct and incorrect predictions of our model, on both shared and target private categories. We include an example for \textit{HMDB$\rightarrow$UCF} (Fig. \ref{fig:output_visualization_hu}) and one for \textit{Epic-Kitchens D1$\rightarrow$D2} (Fig. \ref{fig:output_visualization_ek}). It is possible to observe in Fig. \ref{fig:output_visualization_ek} how the high similarity among distinct \textit{Epic-Kitchens} categories easily leads to incorrect prediction on both shared and unknown classes. On the other hand, it emerges from the example in Fig. \ref{fig:output_visualization_hu} how the model may fail in correctly classifying sequences from \textit{HMDB$\leftrightarrow$UCF}, despite its ability to extract a useful description (e.g. see bottom right example in Fig. \ref{fig:output_visualization_hu}).

\begin{table*}[h!]
    \centering
    \begin{tabular}{|c|c|c|c|}
        \toprule
        \multicolumn{3}{|c|}{\textbf{Ground truth}} & \textbf{Discovered} \\
        \midrule
        turn-on & shake & compress & glass AND brush AND plate AND cup AND fork \\
        drop & knead & scrape & brush AND sponge AND plate AND cup AND fork \\
        grate & extract & crush & chair AND sponge AND plate AND cup AND fork \\
        throw-into & spread & move around & microwave AND brush AND plate AND cup AND fork \\
        turn & throw & remove from & refrigerator AND sponge AND plate AND cup AND fork \\
        see & set & wrap & woman AND carrot AND plate AND cup AND fork \\
        adjust & hang & gather & phone AND sponge AND plate AND cup AND fork \\
        fold & separate & wrap around & brush AND chair AND plate AND cup AND fork \\
        wait-for & flip & press & brush AND chair AND sponge AND cup AND fork \\
        scoop & eat & wrap with & pizza AND glass AND plate AND cup AND fork \\
        taste & heat & rotate & carrot AND brush AND plate AND cup AND fork \\
        drink & wait & fix &mirror AND microwave AND plate AND cup AND fork \\
        turn-off & check & crack & phone AND chair AND sponge AND plate AND cup \\
        drain & look for & read & glass AND chair AND plate AND cup AND fork \\
        squeeze & sprinkle & split & mirror AND sponge AND plate AND cup AND fork \\
        dry & roll & seal & book AND glass AND brush AND chair AND cup \\
        move & peel & press down & cookie AND brush AND sponge AND plate AND fork \\
        empty & unroll & break & book AND woman AND plate AND cup AND fork \\
        unfold & hold & distribute & glass AND chair AND sponge AND plate AND fork \\
        switch-on & spread onto & serve & refrigerator AND chair AND plate AND cup AND fork \\
        put-in & flatten & pat & \\
        spoon & pull down & throw in & \\
        sprinkle-onto & take out & lower & \\
        put-into & remove & take off & \\
        move-into & lift & throw off & \\
        attach-onto & pat down & grind & \\
        twist-off & immerge & spray & \\
        hand & move onto & tap & \\
        \bottomrule
    \end{tabular}
    \caption{List of the actual names of the ground truth target private classes (left) and list of candidate target-private label names identified by \autolabel (right) for the \textit{Epic-Kitchens D1$\rightarrow$D2} setting}
    \label{tab:discovered_ek}
\end{table*}

\end{document}